%% file: neurips_2024.tex
\documentclass{article}

\usepackage[preprint]{neurips_2024}

\usepackage[utf8]{inputenc} 
\usepackage[T1]{fontenc}    
\usepackage{hyperref}       
\usepackage{url}            
\usepackage{booktabs}       
\usepackage{amsfonts}       
\usepackage{nicefrac}       
\usepackage{microtype}      
\usepackage{xcolor}         

\usepackage{amsmath}
\usepackage{amssymb}
\usepackage{mathtools}
\usepackage{amsthm}
\usepackage{makecell}
\usepackage{adjustbox}
\usepackage{multirow}
\usepackage{float}
\usepackage{caption}

\title{FutureNet-LOF: Joint Trajectory Prediction and Lane Occupancy Field Prediction with Future Context Encoding}

\author{%
  Mingkun Wang\\
  Peking University \\
  \And
  Xiaoguang Ren$^*$ \\
  Academy of Military Sciences \\
  \And 
  Ruochun Jin \\
  National University of Defense Technology \\
  \And 
  Minglong Li \\
  National University of Defense Technology \\
  \And
  Xiaochuan Zhang \\
  Academy of Military Sciences \\
  \And 
  Changqian Yu \\
  Meituan \\
  \And 
  Mingxu Wang \\
  Fudan University \\
  \And
  Wenjing Yang\thanks{Corresponding author}\\
  National University of Defense Technology \\
  }

\begin{document}

\maketitle

\begin{abstract}
Most prior motion prediction endeavors in autonomous driving have inadequately encoded future scenarios, leading to predictions that may fail to accurately capture the diverse movements of agents (e.g., vehicles or pedestrians). To address this, we propose FutureNet, which explicitly integrates initially predicted trajectories into the future scenario and further encodes these future contexts to enhance subsequent forecasting.
Additionally, most previous motion forecasting works have focused on predicting independent futures for each agent. However, safe and smooth autonomous driving requires accurately predicting the diverse future behaviors of numerous surrounding agents jointly in complex dynamic environments. Given that all agents occupy certain potential travel spaces and possess lane driving priority, we propose Lane Occupancy Field (LOF), a new representation with lane semantics for motion forecasting in autonomous driving. LOF can simultaneously capture the joint probability distribution of all road participants' future spatial-temporal positions.
Due to the high compatibility between lane occupancy field prediction and trajectory prediction, we propose a novel network with future context encoding for the joint prediction of these two tasks. Our approach ranks 1st on two large-scale motion forecasting benchmarks: Argoverse 1 and Argoverse 2.
\end{abstract}

\section{Introduction}
Essential for safe, effective, and smooth self-driving car planning and control,
predicting future behaviors~\cite{ye2021tpcn,wang2023ganet,wang2023prophnet,cheng2023forecast,liang2020learning,hu2023planning} of road participants is one of the most important and challenging problems in autonomous driving. To address the challenges posed by highly dynamic, volatile movements and complex, diverse driving environments, the research community has made significant efforts to enhance models' predictive capabilities.

Prior works~\cite{wang2023ganet,chai2020multipath,zhou2022hivt,liang2020learning} typically employ well-established encoding networks such as LSTM~\cite{yu2019review}, CNN~\cite{alzubaidi2021review}, Transformer~\cite{han2022survey}, and GNN~\cite{battaglia2018relational} to encode various representations of driving scene information, including rasterized images~\cite{cui2019multimodal}, sparse vectors~\cite{gao2020vectornet,wang2023ganet}, and point clouds~\cite{ye2021tpcn}. After encoding the scene, these methods generally apply simple neural networks to predict multiple possible future trajectories for agents. However, these approaches may encounter mode collapse, where they predict high-frequency patterns from the training data but fail to generate accurate and diverse future predictions.
To capture multiple possible future trajectories, some methods utilize predefined trajectories~\cite{chai2020multipath}, predefined goals~\cite{zhao2021tnt,gu2021densetnt}, or model-based approaches~\cite{liu2021multimodal}, and employ a learned model to score them. However, these methods rely on complex handcrafted preprocessing, resulting in a loss of flexibility to adaptively adjust the trajectories based on current scene information.
Recently, several refinement-based approaches~\cite{wang2023ganet,zhou2023query,shi2024mtr++,shi2022motion,wang2023prophnet} have achieved state-of-the-art performance in motion prediction. 
They first predict preliminary trajectories and then refine them further. 
This refinement process has been proven to effectively promote predictive performance. However, these methods are still insufficient in constructing and encoding future scenarios adequately.

Most existing works~\cite{zhou2022hivt,gao2020vectornet,liang2020learning} tend to prioritize encoding historical observations of the scene while neglecting the modeling of predicted trajectories and the future scenarios where these predicted trajectories lie.
This oversight hampers the effective utilization of map information spanning the past, present, and future.
Observing that many state-of-the-art methods still produce trajectories that veer off road edges, we argue that explicitly encoding the future scenario is crucial for accurate motion prediction. 
Moreover, agents' future motion is highly dependent on future map topology and interactions. Given the high diversity in scene context and agent motion, especially over extended long-term prediction horizons, simply encoding the historically observed scene once is insufficient.

Therefore, we propose FutureNet, which encodes the future scenario by putting the initially predicted trajectories into it. Its core components include a recurrent decoder and a refinement decoder, accompanied by a series of future context encoding modules. In each recurrence or refinement decoding step, we decode the last predicted trajectories and utilize their endpoints as anchors for additional future context encoding.

Additionally, existing prediction methods~\cite{gu2021densetnt,zhao2021tnt,gao2020vectornet} primarily focus on individual or a few key agents, resulting in independent marginal predictions. However, in the context of autonomous driving, such methods struggle to address the formidable challenges of navigating complex and densely populated environments~\cite{ngiam2021scene,shi2024mtr++}. In these environments, autonomous vehicles must consistently assess the behaviors of numerous surrounding agents simultaneously. Therefore, efficient solutions for comprehensive and simultaneous prediction of multiple agents are highly desirable.

Thanks to our proposed FutureNet, we can explicitly construct future scenarios. We observe that the future motion of vehicles is highly constrained by the drivable space of lanes, and agents' future movements result in lane space occupation. Each vehicle occupies a range of potential driving space and holds driving priority within its lane. 
From the perspective of lane driving priority contention, road participants compete for drivable lane space, generating "interaction forces" such as repulsion force for collision avoidance between agents. 
For instance, a vehicle typically avoids encroaching upon the potential driving space of its neighbors, such as cutting in front of a fast-moving car. 
Based on these observations, we propose the concept of Lane Occupancy Field, a novel occupancy field with lane semantics. 
The lane occupancy field samples lane centerlines and lane boundaries to generate map points with specific lane semantics, which are further constructed as an additional layer of dynamic features atop traditional static maps. 
This indicates the probability that the area surrounding the sampled map point is occupied at a specific time step.

The lane occupancy field successfully mitigates the drawbacks of two commonly used representations: trajectory sets~\cite{phan2020covernet,gao2020vectornet,liu2021multimodal,wang2023ganet,varadarajan2022multipath++,ye2021tpcn} and occupancy grids~\cite{mahjourian2022occupancy,casas2021mp3,hong2019rules,kim2022stopnet,agro2023implicit}.
(a) Compared to methods that output trajectory sets, our lane occupancy field has the advantage of capturing the joint probability distribution of future spatial-temporal positions of all road participants simultaneously, rather than the independent marginal probabilities of each agent. It employs a non-parametric representation to output the potential future occupancy spaces for all agents, naturally exhibiting the multi-future characteristics of motion forecasting.
(b) Although occupancy grids can jointly represent the future position distributions of multiple targets, their $H\times W$ image-like representation is computationally intensive and not well compatible with mainstream vector-based motion prediction methods.
In contrast, our lane occupancy field is constructed based on high-definition maps represented in vector form,
which seamlessly integrates with existing vector-based trajectory prediction methods.
Furthermore, compared to occupancy grids, the lane occupancy field inherently carries lane semantics, allowing for the full utilization of known scene priors.

Leveraging the compatibility between lane occupancy field prediction and trajectory prediction, we propose a unified prediction network, termed FutureNet-LOF, for joint lane occupancy field prediction and multi-agent trajectory prediction. This network recurrently advances motion forecasting by utilizing trajectory queries for trajectory decoding and map queries for lane occupancy field prediction.


Our contributions are as follows:
\begin{itemize}
\item We propose FutureNet, a framework for trajectory prediction that emphasizes future context encoding. Our experiments demonstrate the crucial importance of future context encoding for accurate motion prediction.
\item We introduce the concept of the Lane Occupancy Field, a novel representation for motion forecasting in autonomous driving, along with the lane occupancy field prediction task. The LOF approaches the motion prediction problem from the perspective of drivable lanes, opening new avenues for exploration in this domain. This representation addresses several limitations inherent in trajectory sets and occupancy grid representations.

\item We propose the FutureNet-LOF model, which jointly predicts motion trajectories and lane occupancy fields, emphasizing future context encoding through dedicated modules. It excels in both multimodal output and long-term prediction tasks. On one hand, it predicts multiple future trajectories for each agents. On the other hand, it simultaneously provides lane-based motion occupancy for the entire driving scene. 

\item Our method ranks first place on two large-scale motion forecasting benchmarks, Argoverse 1 and Argoverse 2, showcasing its superiority and effectiveness. Additionally, it also achieves state-of-the-art performance on the Argoverse 2 multi-world motion forecasting benchmark.
\end{itemize}

\section{Approach}
\begin{figure}[htbp]
\includegraphics[width=1\textwidth]{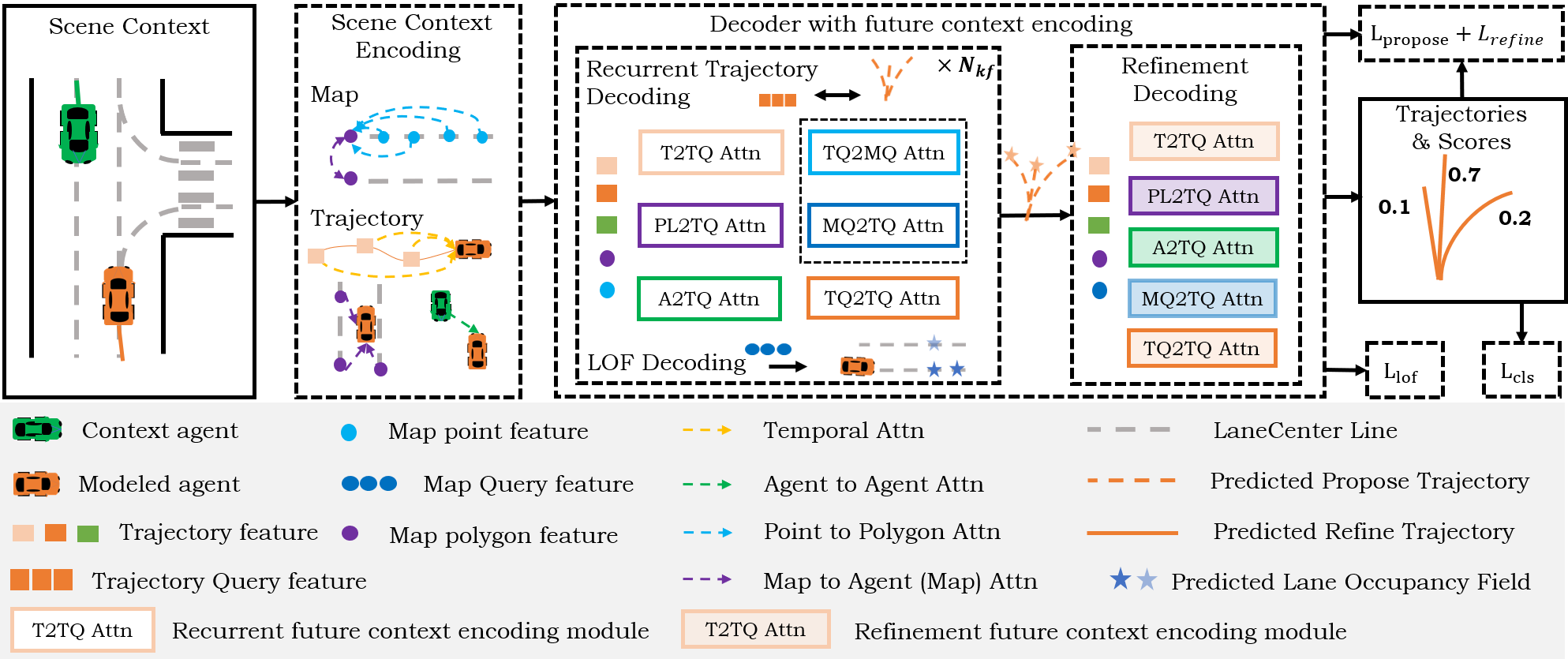}
\centering 
\vspace{-3ex}
\caption{The architecture of our proposed FutureNet-LOF network. }
\label{arch}
\end{figure}

The architecture of our proposed approach is illustrated in Figure~\ref{arch}. We will introduce each module following the data processing pipeline. For an intuitive understanding, please also refer to the unfolded visualization diagram in Appendix~\ref{unfolded}.

\subsection{Problem Formulation}
\textbf{Trajectory Prediction.} 
Following prior works~\cite{liang2020learning, wang2023ganet, zhou2023query}, the scene input is abstracted into a bird's-eye-view (BEV) representation. We employ the prevalent vectorized representation~\cite{gao2020vectornet} for its efficiency and potent capability to capture intricate structured relationships. We denote the historical state of $N_A$ agents as $S_A \in \mathbb{R}^{N_A \times T_h \times C_A}$, where $T_h$ represents the number of historical observation steps, and $C_A$ is the number of feature dimensions of the agent's state (e.g., position, yaw angle, and velocity). The high-definition map provides the map polygons (e.g., lanes and crosswalks), where each map polygon is annotated with sampled points and attributes. We denote the road map as coarse-grained map polygon state $S_M \in \mathbb{R}^{N_M \times C_M}$ and fine-grained map point state $S_m \in \mathbb{R}^{N_m \times C_m}$. Here, $N_M$ indicates the number of map polygons, while $N_m$ is the number of sampled map points. $C_M$ represents the number of attributes for each map polygon, and $C_m$ corresponds to the dimension of the map point feature.
Given map information and historically observed trajectories, we predict $K$ trajectories up to time $T$ and the corresponding confidence scores for each agent.

\textbf{Lane Occupancy Field Prediction.} 
We utilize the sampled points $S_m$ from lane centerlines and boundaries as the basis for the lane occupancy field. In a scenario with $N_m$ lane sampled points, its lane occupancy field $O^{kf}$ at keyframe $kf$ can be represented by an $N_m \times 1$-dimensional vector. The values in this vector range from 0 to 1, indicating the probability of the sampled lane points being occupied by any agent. The ground-truth lane occupancy field labels are generated by computing the distances between agents' future trajectories and lane sampled points at each keyframe. If any agent is within a threshold distance of the sampled point $i$, we set it as occupied, denoted as \( O_i^{kf}=1 \). On both Argoverse 1 and Argoverse 2, we set the distance threshold to two meters and the number of future keyframes to 3. Our lane occupancy field prediction task aims to forecast the occupancy status of these sampled points in the future keyframes.

\subsection{Scene Context Encoding}

\textbf{Multiple parallel local worlds modeling.} 
Inspired by the query-centric modeling approaches in QCNet~\cite{zhou2023query} and MTR++~\cite{shi2024mtr++}, we propose a more comprehensive and intuitive scenario encoding scheme called the multi-parallel local worlds encoding scheme, as shown in Section~\ref{multi_local_worlds}. Specifically, we construct a local spatial-temporal reference coordinate system for all elements in the scene. Each local world has an anchor consisting of coordinates $P$ and orientation $H$ in the global coordinate system, serving as a local world coordinate reference and facilitating positioning in the global context. 
For a polygon (e.g., lane or crosswalk), its entry position and orientation serve as the reference coordinates for the local world. The position and moving direction of each agent state determine its local coordinate system, similar to QCNet~\cite{zhou2023query}. For a map point, its position and lane heading direction establish its local coordinate system. All elements learn a representation by encoding attribute features within their own local world, independent of the global coordinate system.
The attributes $S_A$, $S_m$, and $S_M$ are encoded as follows:
\begin{equation}
    F = MLP(\delta (S))
\end{equation}    
where $\delta (\cdot)$ transforms the features from the global coordinate to an globally independent representation. For example, for each trajectory, we calculate the magnitude of the state displacement at each time step.

At a higher level, we establish $T_h$ trajectory-level local worlds for each trajectory, where each world encompasses trajectories up to the current state, with its reference coordinate being the position and moving direction of the current state. Each map polygon world contains map points sampled on this polygon. To construct features for map polygons and trajectories, we employ a local-world-centric attention mechanism for feature aggregation within these higher-level local worlds.
 
\textbf{Local-world-centric attention.}
Since each local world establishes a globally invariant representation within its scope, we adopt a local-world-centric attention module~\cite{zhou2023query} to model the relationships between different local worlds. Specifically, we apply the attention mechanism to each local world based on the spatial-temporal relative positions of their anchors, as shown in Section~\ref{unfolded}. For example, when considering the $i$-th local world as the query, we transform its surrounding worlds' anchors into the coordinate system of the $i$-th local world. This enables us to determine their relative positions, which include relative distance, relative direction, relative orientation, and time gap. If the time gap is not relevant, such as in the relative position between two map nodes, we set it to zero. Our local-world-centric attention mechanism operates as follows:
\begin{equation}
\label{attn}
 F_i = Attn(Q = F_i, K = [F_j, R_{i,j}]_{ j \in N_i }, V = [F_j,R_{i,j}]_{ j \in N_i }) 
\end{equation}
where $R_{i,j}$ denotes the spatial-temporal relative position embedding between nodes $i$ and $j$, encoding their relative position. $N_i$ represents the neighboring worlds of $i$. For the attention within trajectory or polygon level local world, $N_i$ indicates the states on a same trajectory or map points in a same polygon, while $N_i$ denotes $i$'s neighboring nodes within a certain range between different worlds. 
We utilize local-world-centric attention as a fundamental component for modeling various scene context interactions, including point-to-polygon, agent's temporal attention, polygon-to-polygon, polygon-to-agent, and agent-to-agent attention. Refer to Section~\ref{unfolded} for details.

The encoding network ultimately generates trajectory-level features $F_{A}\in \mathbb{R}^{N_A \times T_h \times D}$, polygon-level features $F_M \in \mathbb{R}^{N_M \times D}$, and point-level features $F_m \in \mathbb{R}^{N_m \times D}$, which are fed into subsequent decoding networks. These features enable a shared scene context for the simultaneous prediction of multiple agents and can also be reused in online prediction, exhibiting translation-rotation invariance and symmetry properties.

\subsection{Decoder with future context encoding}
As shown in Figure~\ref{arch} and Section~\ref{unfolded}, our decoder adopts a recurrent and refinement module but differs from previous methods in a key aspect: future context encoding. 
Most prior refinement-based works~\cite{zhou2023query,shi2024mtr++} primarily focus on historically encoded scene context, failing to fully utilize contextual cues from the future scenario where the predicted trajectories lie. In our FutureNet-LOF, the predicted trajectory is decoded into the global coordinate system and used as an anchor for further future scene encoding after each prediction. We utilize the decoded anchors and the updated features encoded in the future scenario for the next step prediction.
The recurrent decoder consists of two branches: trajectory prediction and lane occupancy field prediction, which interact with each other in the future scenario. Our recurrent decoder recurrently predicts over $N_{kf}$ key steps, forecasting $\frac{T}{N_{kf}}$ waypoints at each recurrent step.

\textbf{Recurrent trajectory decoding.} 
To encode the future scenario, we conduct motion forecasting in a recurrent mode. 
Concretely, we initialize $K$ learnable queries for all agents $TQ^{N_A \times K \times D}$ to decode their trajectories, where each embedding is responsible for decoding a specific trajectory pattern. 
Building upon the concept of multiple parallel local worlds, each query is also regarded as a local world. 
Initially, the trajectory query world's global anchor is set as each agent's current position and orientation. Similar to our scene context encoding step, we transform all other worlds' anchors into the trajectory query local world's coordinate system. Then, we perform temporal-to-trajectory query attention on each trajectory, polygon-to-trajectory query attention, agent-to-trajectory query attention, and attention among each agent's K trajectory queries, based on our local-world-centric attention as in equation~\ref{attn}. Subsequently, we employ an MLP network to decode $\frac{T}{N_{kf}}$ waypoints.

\textbf{Future context encoding.} 
To underscore the significance of future context, we propose constructing a future scenario and encoding this future scenario for further recurrent or refinement prediction. After each prediction, we construct new trajectory query worlds by decoding the endpoints' position and orientation of the last predicted trajectories, which are immersed in the future scenario. 
In a future scenario, the new trajectory query local worlds will occupy the drivable space of maps while being constrained by the maps and future interactions between agents. As significant changes may occur in the surrounding environment when predicting over longer horizons, re-encoding the future scene context becomes crucial.
Therefore, we re-encode the context within this future scenario. Our future context encoding modules consist of a series of local-world-centric attentions, each focused on encoding a specific type of interaction. Specifically, we conduct attention similar to the first recurrent trajectory decoding step to encode new trajectory query worlds based on the local worlds' relative positions. Then, an MLP network is utilized to decode trajectories based on the updated trajectory query features to advance prediction. This process continues recurrently until the complete proposed trajectories are predicted.

\textbf{Lane occupancy field decoding.} 
Coupled with each iteration of the trajectory prediction loop, we perform lane occupancy field prediction at each keyframe. We construct map query local worlds $MQ^{N_m \times D}$, employing the features of sampled map points as learnable queries and using the sampled points' positions and orientations as anchors.
Considering that trajectories occupy map space in a future scenario, activating our lane occupancy field, we perform trajectory query-to-map query attention centered on map queries. Additionally, we employ attention from map queries to trajectory queries to capture the influence of the lane occupancy fields on trajectory query worlds. 
After obtaining updated map query features, we utilize an MLP network to decode the logits for each map query. Subsequently, we apply a sigmoid function to derive the lane occupancy fields.

\begin{table}[htbp]
\caption{Quantitative results on the Argoverse 2 motion forecasting leaderboard.}
\label{test}
\begin{adjustbox}{width=\textwidth}
\centering
\begin{tabular}{c|cccc|ccc}
\toprule
Method    & b-minFDE$_6$ $\downarrow$	 &minFDE$_6$ $\downarrow$	&minADE$_6$ $\downarrow$	&MR$_6$ $\downarrow$	&minFDE$_1$ $\downarrow$	& minADE$_1$ $\downarrow$	&MR$_1$ $\downarrow$ \\

\midrule

FRM~\cite{park2023leveraging}	&2.47	&1.81	&0.89	&0.29	&5.93	&2.37	&0.71   \\
HOME \& GOHOME~\cite{gilles2022gohome}	&2.16	&1.51	&0.88	&0.20	&4.71	&1.95	&0.64  \\
HPTR~\cite{zhang2023hptr} &2.03	&1.43	&0.73	&0.19	&4.61	&1.84	&0.61   \\
Gorela~\cite{cui2023gorela}	&2.01	&1.48	&0.76	&0.22	&4.62	&1.82	&0.61 \\
MTR~\cite{shi2022motion}	&1.98	&1.44	&0.73	&0.15	&4.39	&1.74	&0.58   \\
GANet~\cite{wang2023ganet}	&1.96	&1.34	&0.72	&0.17	&4.48	&1.77	&0.59   \\
Forecast-MAE~\cite{cheng2023forecast}	&1.91	&1.34	&0.69	&0.17	&4.15	&1.66	&0.59   \\
TENET~\cite{feng2023macformer} &1.90	&1.38	&0.70	&0.19	&4.69	&1.84	&0.61   \\
HeteroGCN~\cite{gao2023dynamic} 	&1.90	&1.34	&0.69	&0.18	&4.40	&1.72	&0.59    \\
ProphNet~\cite{wang2023prophnet}	&1.88	&1.32	&0.66	&0.18	&4.77	&1.76	&0.61    \\
QCNet~\cite{zhou2023query}	&1.78	&1.19	&0.62	&0.14	&3.96	&1.56	&0.55  \\

\textbf{FutureNet-LOF (Ours)}	& \textbf{1.63}	& \textbf{1.07}	& \textbf{0.58}	& \textbf{0.12}	& \textbf{3.63}	&\textbf{1.46}	&\textbf{0.51}    \\
\bottomrule
\end{tabular}
\end{adjustbox}
\vspace{-2ex}
\end{table}

\textbf{Trajectory refinement decoding.} 
To further enhance the encoding of future context and refine the predicted proposal trajectories, we construct a trajectory refinement decoding module. Similarly, we treat each refinement trajectory query as a local world, with the anchor being the endpoint position and orientation of the previously predicted complete trajectory.
To construct features for the refinement query, we employ a GRU network to encode the predicted trajectory. Next, we encode the refinement trajectory query world using local-world-centric attention, considering the impact of the agent's temporal observed motion, the influence of map polygons, and the interactions between agents. We also consider the influence of the map query worlds on the refinement trajectory queries. Finally, we facilitate interactions between each agent's $K$ refinement trajectory query worlds, enabling communication between the parallel worlds. After obtaining updated query features, we use an MLP network to decode the offset relative to the initially predicted proposal trajectory and the confidence score $p$ for each trajectory.

Following the approach in~\cite{zhou2022hivt,zhou2023query}, we represent the distribution of predicted trajectories using a mixture of Laplace distributions. The predicted position distribution of the $i$-th agent can be formulated as:
\begin{equation}
\label{dis}
    f(s_i) = \sum_{k=1}^Kp_i^k\prod_{t=1}^T\cdot Laplace(s_{i}^{k,t}|\mu_{i}^{k,t},b_{i}^{k,t})
\end{equation}

where $p_i^k$ is the mixing coefficient, and the k-th mixture component’s Laplace density at time step t is parameterized by the location $\mu_{i}^{k,t}$ and the scale $b_{i}^{k,t}$. The predicted trajectories can be obtained by directly extracting the centers of the predicted Laplace components.

\subsection{Training objective functions} 
Our framework employs supervised end-to-end training, incorporating lane occupancy field loss, regression loss, and classification loss. The lane occupancy field loss is a balanced binary logistic cross-entropy between predicted and ground-truth fields. Notably, unoccupied lane fields significantly predominate in the ground-truth samples. To address this issue, we use a balanced binary cross-entropy loss BCE with a positive class weight $\alpha=0.8$. The losses are aggregated across all lane points per time step, denoted as:
\begin{equation}
    L_{lof}=\frac{1}{N_{kf} \times N_m}\sum_{kf=1}^{N_{kf}}\sum_{i=1}^{N_m}BCE(O_i^{kf},\Tilde{O}_i^{kf})
\end{equation}

With only one ground-truth trajectory available in trajectory prediction, we employ a winner-takes-all~\cite{lee2016stochastic} strategy. The regression loss is specifically targeted at minimizing the negative log-likelihood for the best-predicted proposed trajectory and its refined trajectory. Additionally, we utilize a classification loss $L_{cls}$ to optimize the mixing coefficients predicted by the refinement module. This loss aims to minimize the negative log-likelihood of Equation ~\ref{dis}.
The final loss is defined as:
\begin{equation}
    L=L_{propose}+L_{refine}+\beta L_{cls}+ \rho L_{lof}
\end{equation}

\section{Experiments}
\subsection{Experimental Settings}
\textbf{Datasets.}
We evaluate our approach on two large-scale motion forecasting datasets, Argoverse 1~\cite{chang2019argoverse} and Argoverse 2~\cite{wilson2023argoverse}, which offer a wide range of real-world driving scenarios. The Argoverse 1 dataset comprises over 300K scenarios from Pittsburgh and Miami, each lasting 5 seconds. For the test set, only the initial 2-second trajectories are provided, necessitating us to predict and submit the subsequent 3 seconds for evaluation on the official benchmark.
The Argoverse 2 dataset encompasses 250K scenarios spanning over 2,000 kilometers across six geographically diverse cities. This dataset showcases advancements in scenario diversity, data quality, agent categories, and prediction horizon. Each scenario spans 11 seconds, during which we observe 5 seconds and predict the subsequent 6 seconds.

\textbf{Metrics.}
We utilize widely adopted evaluation metrics~\cite{chang2019argoverse} minADE$_K$, minFDE$_K$, b-minFDE$_K$, and MR$_K$. The metric minADE$_K$ calculates the average L$_2$ distance in meters between the ground-truth trajectory and the best of $K$ predicted trajectories across all time steps. The metric minFDE$_K$ is defined as the L$_2$ displacement error between ground-truth trajectories and the best-predicted trajectories at the final time step. The b-minFDE$_K$ also considers the trajectory's confidence. The MR$_K$ is the ratio of scenarios where the predicted trajectory exceeds 2 meters of the ground truth according to the FDE.
In lane occupancy field prediction, we employ Area under the Curve (AUC) and Intersection over Union (IoU). AUC uses a linearly-spaced set of thresholds in $[0, 1]$ to compute pairs of precision and recall values and estimate the area under the PR-curve. IoU measures the overlap between the predicted lane occupancy field and ground-truth lane occupancy field as:
\begin{equation}
 IoU_{th}(O^{kf},\Tilde{O}^{kf}) =\frac{\sum^{N_m}_{i=1} 1[O_i^{kf} > th] \cdot \Tilde{O}_i^{kf}}{\sum^{N_m}_{i=1}(O_i^{kf} + \Tilde{O}_i^{kf}-O_i^{kf} \cdot \Tilde{O}_i^{kf})}  
\end{equation}
where $1[\cdot]$ equals 1 when the predicted occupancy probability is greater than the threshold $th$.

\begin{table}[tbp]\small
\caption{Lane occupancy field prediction results, based on the Argoverse2 validation set.}
\label{lof}
\centering
\begin{tabular}{c|c|cccccc}
\toprule
\multicolumn{2}{c}{\multirow{2}{*}{Method}}&	\multicolumn{3}{c}{IoU $\uparrow$}	&\multicolumn{3}{c}{AUC $\uparrow$} \\
\cline{3-8}
\multicolumn{2}{c}{}  &2s	&4s	&6s	&2s	&4s	&6s \\
\midrule
Trajectory-based method	&1	&0.41	&0.29	&0.22&	0.73	&0.74	&0.71  \\
with render thresholds (m)	&2	&0.58	&0.32	&0.23	&0.97	&0.93	&0.90  \\
	&3	&0.33	&0.20	&0.15	&0.96	&0.93	&0.91   \\
	&4	&0.21	&0.14	&0.11	&0.93	&0.90	&0.90    \\

 \midrule
FutureNet-LOF	&0.5	&0.66	&0.48	&0.39	  \\
with overlap thresholds	&0.7	&\textbf{0.70}	&\textbf{0.51}	&\textbf{0.40}	&\textbf{0.98}   &\textbf{0.97}  &\textbf{0.95}  \\
	&0.9	&0.64	&0.39	&0.29	  \\
\bottomrule
\end{tabular}
\vspace{-2ex}
\end{table}

\subsection{Main Results}
For trajectory prediction, we compare our approach with state-of-the-art methods on the Argoverse 1 and Argoverse 2 motion forecasting benchmarks. As shown in Table ~\ref{test} and  ~\ref{test1}, our method ranks 1st on the leaderboards of both Argoverse 1 and Argoverse 2, noting that our FutureNet-LOF method outperforms all existing methods across all evaluation metrics on the more challenging Argoverse 2 dataset. It is also the champion approach of the CVPR 2024 Argoverse 2 motion forecasting challenge.
Despite the performance of existing methods approaching saturation on the Argoverse 1 dataset, our approach further pushes the performance boundaries, which demonstrates our approach's generalization ability. Our method has also achieved state-of-the-art performance on the Argoverse 2 multi-world motion forecasting benchmark. As shown in Figure~\ref{test_multi_world}, our results on the official leaderboard significantly surpass FJMP~\cite{rowe2023fjmp}, a motion prediction method with multi-agent interaction design. In the key performance metrics avgMinFDE6 and avgMinADE6, we achieved improvements of 33.9\% and 28.4\% over FJMP, respectively.

To validate the effectiveness of our lane occupancy field prediction model, we predict the status of lane occupancy fields at keyframes 2s, 4s, and 6s on the Argoverse 2 validation set. As a comparative baseline, we use trajectories predicted by the trajectory prediction branch of the FutureNet-LOF model to render lane occupancy field at each keyframe. Similar to the ground-truth generation process, if the predicted trajectories fall within a certain distance threshold of lane points, the lane occupancy field is set to 1. We use four different distance threshold values: 1 meter, 2 meters, 3 meters, and 4 meters for rendering. When computing the IoU metric for our lane occupancy prediction model, we set overlap thresholds of 0.5, 0.7, and 0.9. If the predicted probability exceeds the threshold, the lane occupancy field is set to 1; otherwise, it is set to 0. As shown in Table~\ref{lof}, it is evident that our lane occupancy field prediction model consistently outperforms all trajectory rendering-based lane occupancy fields in terms of both IoU and AUC metrics, validating the effectiveness of our lane occupancy field prediction model.

\begin{table}[tbp]\small
\caption{Ablation studies on the components of the decoder, based on the Argoverse 2 validation set.}
\label{ablation}
\begin{adjustbox}{width=\textwidth}
\centering
\begin{tabular}{@{}cccc|cccc@{}}
\toprule
\vspace{-5px}
\multirow{3}{*}{One step} & \multirow{3}{*}{Recurrence} & \multirow{3}{*}{Refinement} & \multirow{3}{*}{LOF} & \multirow{3}{*}{\makecell[c]{b-minFDE$_6$$\downarrow$}} & \makecell[c]{minFDE$_6$$\downarrow$} & \makecell[c]{minADE$_6$$\downarrow$} & \makecell[c]{MR$_6$$\downarrow$} \\
\vspace{-4px}
&&&& & ------------ & ------------ & ------------ \\
&&&&& \makecell[c]{minFDE$_1$$\downarrow$} & \makecell[c]{minADE$_1$$\downarrow$} & \makecell[c]{MR$_1$$\downarrow$} \\

\midrule
\vspace{-5px}
\multirow{3}{*}{\checkmark} & \multirow{3}{*}{} & \multirow{3}{*}{} & \multirow{3}{*}{} & \multirow{3}{*}{1.92} & 1.36 & 0.74 & 0.18  \\
\vspace{-4px}
&&&&& ------ & ------ & ------  \\
&&&& &4.09 &1.62 &0.59  \\
\cline{5-8}
\vspace{-5px}
\multirow{3}{*}{} & \multirow{3}{*}{\checkmark} & \multirow{3}{*}{} & \multirow{3}{*}{} & \multirow{3}{*}{1.82} & 1.21 & \textbf{0.71} & 0.15  \\
\vspace{-4px}
&&&&& ------ & ------ & ------  \\
&&&&&4.02 &1.61 &0.57  \\
\cline{5-8}
\vspace{-5px}
\multirow{3}{*}{\checkmark} & \multirow{3}{*}{} & \multirow{3}{*}{\checkmark} & \multirow{3}{*}{} & \multirow{3}{*}{1.82} & 1.22 & 0.72 & 0.15  \\
\vspace{-4px}
&&&&& ------ & ------ & ------  \\
&&&&&4.05 &1.61 &0.57  \\
\cline{5-8}

\vspace{-5px}
\multirow{3}{*}{} & \multirow{3}{*}{\checkmark} & \multirow{3}{*}{\checkmark} & \multirow{3}{*}{} & \multirow{3}{*}{1.80} & 1.19 & \textbf{0.71} & 0.14  \\
\vspace{-4px}
&&&&& ------ & ------ & ------  \\
&&&& &4.06 &1.62 &0.57  \\
\cline{5-8}

\vspace{-5px}
\multirow{3}{*}{} & \multirow{3}{*}{\checkmark} & \multirow{3}{*}{\checkmark} & \multirow{3}{*}{\checkmark} & \multirow{3}{*}{\textbf{1.79}} & \textbf{1.16} & \textbf{0.71} & \textbf{0.13}  \\
\vspace{-4px}
&&&&& ------ & ------ & ------  \\
&&&&&\textbf{3.97} &\textbf{1.59} &\textbf{0.55}  \\

\bottomrule
\end{tabular}
\end{adjustbox}
\vspace{-3ex}
\end{table}

\begin{table}[tbp]
\caption{Ablation studies of the future context encoding modules on the Argoverse 2 validation set.}
\label{ablation_module}
\begin{adjustbox}{width=\textwidth}
\centering
\begin{tabular}{@{}cccc|cccc@{}}
\toprule
\vspace{-5px}
\multirow{3}{*}{Temporal} & \multirow{3}{*}{Map Polygon} & \multirow{3}{*}{Agent Social} & \multirow{3}{*}{Mode} & \multirow{3}{*}{\makecell[c]{b-minFDE$_6$$\downarrow$}} & \makecell[c]{minFDE$_6$$\downarrow$} & \makecell[c]{minADE$_6$$\downarrow$} & \makecell[c]{MR$_6$$\downarrow$} \\
\vspace{-4px}
&&&& & ------------ & ------------ & ------------ \\
&&&&& \makecell[c]{minFDE$_1$$\downarrow$} & \makecell[c]{minADE$_1$$\downarrow$} & \makecell[c]{MR$_1$$\downarrow$} \\

\midrule
\vspace{-5px}
\multirow{3}{*}{\checkmark} & \multirow{3}{*}{} & \multirow{3}{*}{\checkmark} & \multirow{3}{*}{\checkmark} & \multirow{3}{*}{1.96} & 1.31 & 0.73 & 0.17  \\
\vspace{-4px}
&&&&& ------ & ------ & ------  \\
&&&&&4.93 &1.88 &0.64  \\
\cline{5-8}
\vspace{-5px}
\multirow{3}{*}{\checkmark} & \multirow{3}{*}{\checkmark} & \multirow{3}{*}{} & \multirow{3}{*}{\checkmark} & \multirow{3}{*}{1.83} & 1.21 & \textbf{0.71} & 0.15  \\
\vspace{-4px}
&&&&& ------ & ------ & ------  \\
&&&&&4.31 &1.69 &0.58  \\
\cline{5-8}
\vspace{-5px}
\multirow{3}{*}{\checkmark} & \multirow{3}{*}{\checkmark} & \multirow{3}{*}{\checkmark} & \multirow{3}{*}{} & \multirow{3}{*}{1.81} & 1.21 & \textbf{0.71} & 0.15  \\
\vspace{-4px}
&&&&& ------ & ------ & ------  \\
&&&&&4.10 &\textbf{1.64} &\textbf{0.57}  \\
\cline{5-8}

\vspace{-5px}
\multirow{3}{*}{\checkmark} & \multirow{3}{*}{\checkmark} & \multirow{3}{*}{\checkmark} & \multirow{3}{*}{\checkmark} & \multirow{3}{*}{\textbf{1.80}} & \textbf{1.19} & \textbf{0.71} & \textbf{0.14}  \\
\vspace{-4px}
&&&&& ------ & ------ & ------  \\
&&&& &\textbf{4.06} &\textbf{1.62} &\textbf{0.57}  \\

\bottomrule
\end{tabular}
\end{adjustbox}
\vspace{-3ex}
\end{table}

\subsection{Ablation Study}
We conduct ablation study on the validation dataset of Argoverse 2 to investigate the effectiveness of each component, as presented in Table~\ref{ablation} and Table~\ref{ablation_horizen}. Without future context encoding, the first row represents our baseline model, where we predict complete trajectories at one shot after trajectory query encoding.

\textbf{Effects of the recurrence with future encoding.} Compared to the baseline method, our FutureNet models decode trajectories and encode future context, and then advance further prediction in the second column with a recurrence equals 3. The experimental results in the first two rows of Table~\ref{ablation} demonstrate that our approach of encoding future context significantly improves trajectory prediction performance, improving MR$_6$ by more than 16\% and minFDE$_6$ by more than 11\%. This verifies the importance of our future context encoding module.

\textbf{Effects of the refinement with future encoding.} As shown in the third row of Table~\ref{ablation}, our FutureNet with trajectory refinement, which utilizes future context encoding, also significantly enhances trajectory prediction performance. Moreover, integrating both recurrence and refinement models results in superior performance, further validating the importance of encoding future context.

\textbf{Effects of the lane occupancy prediction branch.} We integrate the tasks of lane occupancy field prediction and trajectory prediction into a unified framework as FutureNet-LOF model. It predicts multi-agent trajectories while also outputting lane occupancy fields for the entire scene. Incorporating the LOF prediction branch has further enhanced our trajectory prediction performance.

\textbf{Effects of different future context encoding modules.} We study the impact of different future encoding modules, with the results shown in Table~\ref{ablation_module}. When we encode the map data only during the initial prediction and do not re-encode it in future scenarios during recurrent and refinement predictions, the prediction performance significantly deteriorates. Incorporating the future map encoding module improves the prediction performance by 9.2\%, 17.6\%, 17.6\%, and 13.8\% in minFDE6, MR6, minFDE1, and minADE1, respectively. Additionally, the future scenario encoding module, which includes social attention and mode attention, further enhances the model's prediction performance. The combination of these modules achieves the best overall performance.

\begin{figure}[tbp]
\includegraphics[width=1\textwidth]{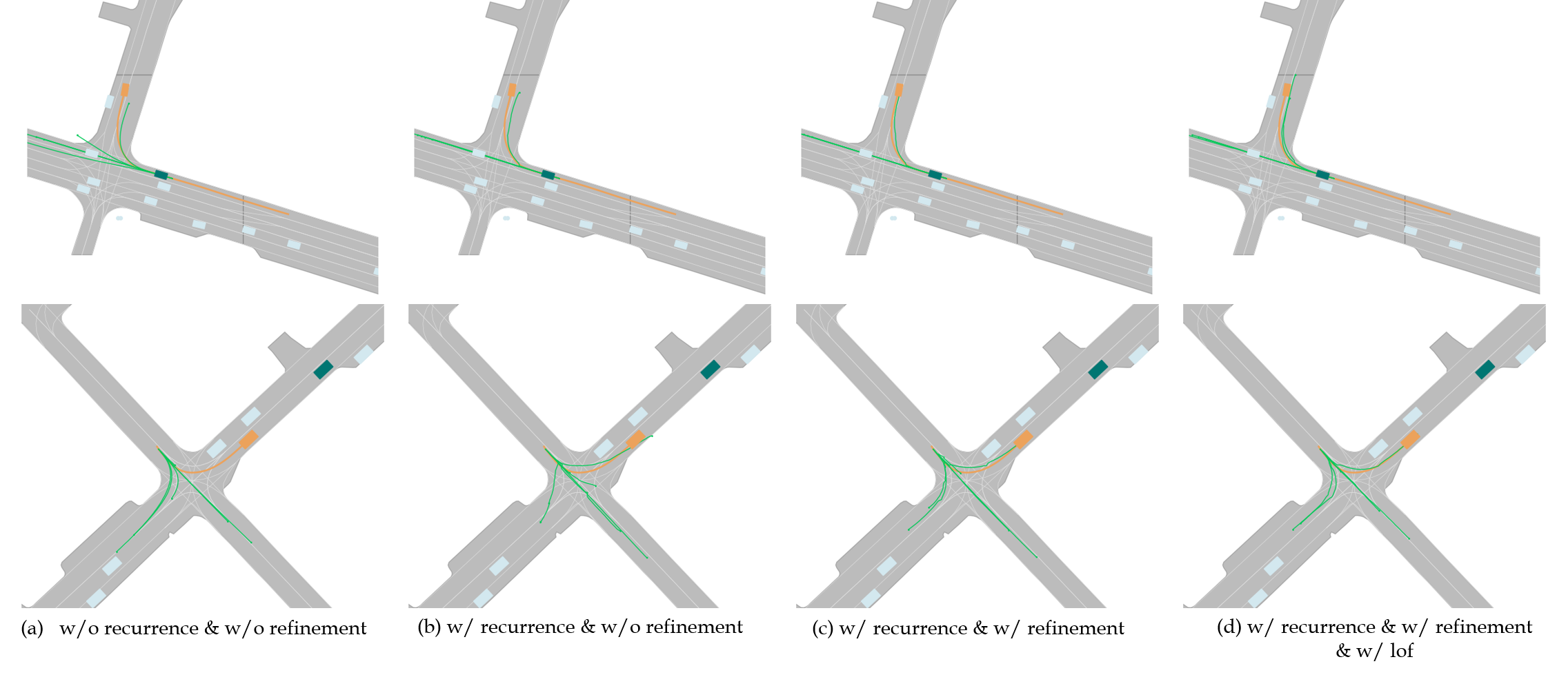}
\centering 
\vspace{-4ex}
\caption{Trajectory prediction qualitative results on the Argoverse 2 validation set. The self-driving car is depicted by a green bounding box, while the focal agent's box and ground-truth trajectories are displayed in orange. Predicted trajectories are shown in green.}
\label{traj_vis}
\end{figure}

\begin{figure}[tbp]
\includegraphics[width=1\textwidth]{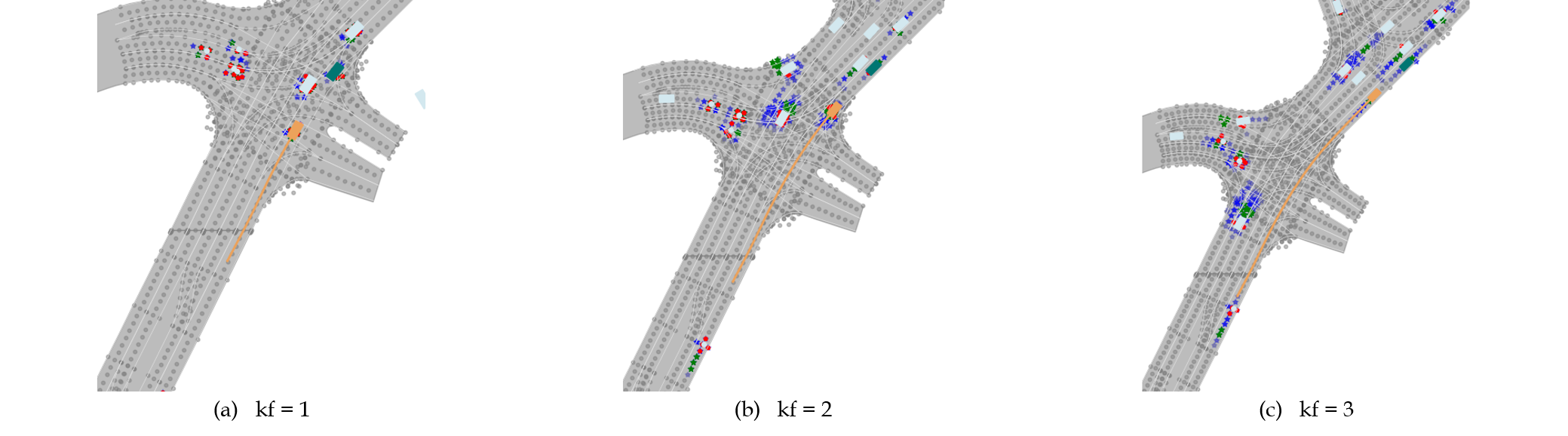}
\centering 
\vspace{-4ex}
\caption{Lane occupancy field prediction qualitative results on the Argoverse 2 validation set. The ground-truth lane occupancy fields are displayed in red. Our predicted fields are shown in blue. The green stars represent lane occupancy field rendered based on our predicted trajectories adopting a distance threshold of 2 meters. }
\label{lof_vis}
\end{figure}

\subsection{Qualitative Results}

We conducted qualitative analysis by visualizing the predicted results. As depicted in Figure~\ref{traj_vis}, the first scenario involves vehicles traversing a crossroad with multiple possible futures. Trajectory predictions from the model without recurrent future encoding generate trajectories that deviate from road topology, veering off the road edges. Models with recurrent future context encoding correct these errors. Trajectories with refinement are smoother and more accurate. Notably, our model incorporating the LOF prediction branch also generates reasonable and accurate predictions, predicting two trajectories for the right-turn mode, which align better with the ground-truth motion.

The second scenario involves vehicles starting from a stationary status in front of a crossroad, where there is insufficient motion history to anticipate the agents’ intentions.. Models without future encoding fail to capture the agents' true motion, missing the left-turn mode. Models with recurrent future encoding capture this pattern, while models with global refinement and lane occupancy field supervision produce more reasonable and accurate trajectories.

We visualize our model's lane occupancy field prediction results. As shown in Figure~\ref{lof_vis}, compared to the ground truth, the predicted lane occupancy fields of FutureNet-LOF are accurate, effectively depicting lane occupancy situations in driving scenarios and significantly outperforming lane occupancy fields rendered based on predicted trajectories.

\section{Conclusion}
In this paper, we introduce a new motion prediction representation called lane occupancy field. We propose a FutureNet framework for trajectory prediction, validating the crucial role of the future context encoding. Additionally, we propose a motion forecasting network, FutureNet-LOF, which jointly predicts lane occupancy fields for the lane space and trajectories for agents. Our method demonstrates outstanding performance on large-scale motion forecasting datasets.


\include{references}

\newpage
\appendix
\section{Related Work}
\subsection{Scene representation}
Given a variety of heterogeneous inputs~\cite{wang2023ganet, wilson2023argoverse,ettinger2021large}, such as the high-definition maps and the past trajectories of agents, the motion forecasting problem requires learning rich representations from the traffic scene. Extensive works explore different representations, including rasterized image~\cite{hong2019rules,chai2020multipath,phan2020covernet,tang2019multiple}, vector-based representation~\cite{wang2023ganet,liang2020learning,zhou2023query,zhou2022hivt,shi2024mtr++}, and point cloud~\cite{ye2021tpcn} representation to encode the scene context. 

Early works render driving scene information into 2D bird's-eye view images~\cite{hong2019rules,chai2020multipath,phan2020covernet,tang2019multiple}, with different elements occupying distinct image channels. These approaches allow for the direct use of well-established convolutional neural networks (CNNs) to encode scene context. However, these 2D image-based representations have inherent drawbacks: information loss during the rendering process, limited receptive fields of convolutional networks that hinder capturing complex dependencies in driving scenes, such as road topology, and the computational burden of CNN-based modeling methods.
IntentNet~\cite{casas2018intentnet} develops a multi-task model that uses a convolutional network-based detector to extract features from rasterized maps. Similarly, MultiPath~\cite{chai2020multipath} utilizes a scene convolutional neural network to extract scene features, encoding the states and interactions of participants within a top-down scene representation.

Recently, the research community has shifted towards encoding schemes based on sparse vector representations~\cite{wang2023ganet,liang2020learning,zhou2023query,zhou2022hivt,shi2024mtr++} due to their computational efficiency and ability to capture complex interactions. Graph-based methods construct graph-structured representations from high-definition map inputs, preserving lane connectivity. By using permutation-invariant set operations, such as pooling, graph convolutions, and attention mechanisms, vector-based approaches can efficiently aggregate sparse information in driving scenes.
VectorNet~\cite{gao2020vectornet} represents map elements and agent trajectories as polylines, then uses a global interaction graph to integrate features from both maps and agents. LaneGCN~\cite{liang2020learning} constructs a graph of map nodes and introduces a novel graph convolution. LaneRCNN~\cite{zeng2021lanercnn} learns a graph-based representation for each agent, encoding its past motion and local map topology, and then uses a shared global lane graph to capture interactions between agents.
Due to the advantages of sparse vector representations, our method also adopts this approach.

Additionally, some studies employ point cloud-based learning methods for motion prediction. These methods represent participants' trajectories and maps using sampled points, offering simplicity and robustness. TPCN~\cite{ye2021tpcn} treats each agent as an unordered point set and applies point cloud learning models to encode scene information.

\subsection{Scene context normalization}
Scene normalization~\cite{ngiam2021scene} is a crucial aspect of scene context encoding, with two mainstream approaches currently prevalent: agent-centric modeling~\cite{zhou2022hivt,gu2021densetnt,gao2020vectornet,zhao2021tnt} and whole-scene modeling~\cite{wang2023ganet,phan2020covernet,biktairov2020prank}. 

Numerous methods~\cite{zhou2022hivt,gu2021densetnt,gao2020vectornet,zhao2021tnt} adopt the agent-centric encoding method, which encodes scene elements centered on each agent, demonstrating commendable accuracy. 
This approach typically normalizes all scene elements based on the current position and orientation of the focal agent. Such normalization method enhances training stability and demonstrates excellent prediction accuracy. However, agent-centric normalization requires re-normalizing the entire scene for each predicted agent, necessitating the re-encoding of scene information. Additionally, whenever the observation window moves forward, all scene elements need to be re-normalized and re-encoded based on the agent's current state. Consequently, the computational cost of these methods grows linearly with the number of focal agents, leading to scalability challenges in crowded urban environments.

Another approach~\cite{wang2023ganet,phan2020covernet,biktairov2020prank}, the whole-scene shared encoding method, typically normalizes all scene elements based on the current state of a focal agent or the self-driving car. Unlike the agent-centric normalization approach, this method allows all agents in the scene to share the same coordinate system. This reduces the computational overhead of re-normalizing and re-encoding the scene for each focal agent that needs prediction. However, this shared normalization method can also diminish the accuracy of predictions for non-central agents, as the distribution of shared scene information is uneven for non-centric agents.

Recently, a class of query-centric methods~\cite{zhou2023query,shi2024mtr++} has emerged, which encode scene elements in a local reference frame and capture interactions between elements based on their relative positions. This approach generates globally invariant feature representations that facilitate feature reuse. However, these methods' designs lack comprehensiveness and intuitiveness. Inspired by recent advancements in query-centric approaches, we introduce a more comprehensive and intuitive scheme named multi-parallel local world encoding, similar to query-centric modeling.

\subsection{Interaction modeling}
Interaction plays a pivotal role in motion prediction. Early methods in motion prediction primarily emphasize motion and interaction modeling, seeking to elucidate the intricate movements of agents by probing potential "interactions." Traditional approaches like the Social Force Model~\cite{helbing1995social} utilize handcrafted features and rules to delineate interactions and constraints among agents. Subsequently, deep learning methods revolutionize this task with significant performance enhancements. Methods such as Social LSTM~\cite{alahi2016social} and SR-LSTM~\cite{zhang2019sr} employ LSTM variants to implicitly capture interactions. Some methods employ graph networks~\cite{battaglia2018relational} to construct interactions between agents. For instance, LaneGCN~\cite{liang2020learning} utilizes a multi-step graph network approach to establish interactions between maps. GNN-TP~\cite{wang2019unsupervised} introduces a graph neural network methodology for interaction inference and trajectory prediction. 
Presently, advanced motion prediction methods leverage Transformer~\cite{wang2023ganet,shi2022motion,shi2024mtr++,zhou2023query,zhou2022hivt,mercat2020multi,wang2023prophnet} architectures to fuse agent motion histories, maps, and their interactions. Typically, they construct fully connected graphs within agent neighborhoods and employ attention mechanisms for information fusion.

Previous works~\cite{liang2020learning} primarily focus on encoding interactions from the observed history, but recent studies begin to consider interactions in future scenarios~\cite{wang2023ganet,shi2022motion}. For example, GANet~\cite{wang2023ganet} predicts potential future endpoint regions for agents and aggregates interaction features between the map and agents within these regions. However, these initial attempts do not fully exploit the potential of future scene encoding. Our approach fully embraces future scenes, providing comprehensive encoding to enhance trajectory prediction. To capture the intricate interactions between agents and maps, we employ a factorized attention method, encompassing map polygon-to-map polygon attention, map polygon-to-agent attention, and so on, to effectively model rich interaction information.

\subsection{Multi-future prediction}
Multi-future prediction is crucial in motion forecasting due to the highly complex and diverse nature of object movements in driving scenarios~\cite{wang2023ganet,liu2021multimodal}. Effectively capturing all possible future motions is essential for ensuring the safe planning and control of autonomous vehicles. Therefore, multi-future prediction has become a vital aspect of motion forecasting. In trajectory prediction, methods typically address this uncertainty by outputting multiple trajectories along with their associated confidence scores.

Some approaches~\cite{wang2023ganet,gao2020vectornet} employ simple neural networks, such as multilayer perceptron (MLP), long short-term memory (LSTM), gated recurrent unit (GRU), and convolutional neural network (CNN), to predict multiple potential future trajectories directly from the extracted agent features after scene encoding. However, these methods face significant challenges in multi-future prediction due to the training data containing only one ground truth trajectory. They may suffer from mode collapse, wherein the models predict high-frequency patterns from the training data but fail to generate accurate and diverse future predictions based on specific scene information.
To mitigate this issue, prior studies~\cite{liang2020learning, ye2023bootstrap} have proposed various strategies. For example, some approaches use a winner-takes-all training strategy, where loss computation and backpropagation are performed only on the best-predicted mode~\cite{wang2023ganet,zhou2023query,zhou2022hivt}, encouraging each prediction branch to capture a distinct mode. Some works adopt model ensembling~\cite{ye2023bootstrap,chai2020multipath}, which aggregates predictions from multiple models to enhance diversity and accuracy.

Generative methods sample from latent variables to introduce randomness for multi-future prediction. Examples include approaches based on Variational Autoencoders (VAEs)~\cite{lee2017desire} and Generative Adversarial Networks (GANs)~\cite{gupta2018social}. However, each prediction using generative methods requires multiple independent samplings and forward propagations, which can not guarantee sample diversity. Additionally, due to the inherent randomness and interpretability challenges of generative methods, they have not yet been widely applied in autonomous driving systems.

Anchor-based approaches have demonstrated excellent performance in multi-future prediction. These methods use preprocessed trajectory anchors~\cite{tang2019multiple, phan2020covernet} or candidate goals~\cite{zhao2021tnt,gu2021densetnt} sampled from the map as priors to guide predictions. CoverNet~\cite{phan2020covernet} and MultiPath~\cite{tang2019multiple} incorporate predefined anchor trajectories as prior knowledge to identify different modes and mitigate the risk of mode collapse. Recently, goal-based prediction methods~\cite{zhao2021tnt,gu2021densetnt} have also proven effective. TNT~\cite{zhao2021tnt} samples dense goal candidates along the lane and generates trajectories associated with high-scoring goals. LaneRCNN~\cite{zeng2021lanercnn} treats each lane segment as an anchor, while GoalNet~\cite{zhang2021map}identifies possible goals and employs a prediction head for each goal. DenseTNT~\cite{gu2021densetnt} introduces a trajectory prediction model that outputs a set of trajectories from densely sampled goal candidates.
However, the superior performance of these methods depends on the quality of predefined anchors~\cite{wang2023ganet,shi2024mtr++}, which lack adaptability. 

Additionally, some methods based on predefined models, such as mmTransformer~\cite{liu2021multimodal}, utilize a region-based training strategy to ensure that each prediction head captures specific motion patterns. Heatmap-based methods~\cite{gilles2022gohome,naser2020heat} use heatmap outputs to naturally represent the distribution of multiple future trajectories. The HOME method~\cite{naser2020heat} predicts a heatmap of future probability distributions and employs a deterministic sampling algorithm to optimize and output the predicted trajectories. GOHOME~\cite{gilles2022gohome} combines a vector-based scene representation with a heatmap output representation, enhancing the prediction process.

Recently, a class of refinement methods~\cite{wang2023ganet,zhou2023query,shi2022motion,shi2024mtr++,wang2023prophnet} has emerged that first make an initial trajectory prediction, treat it as an anchor, and then further refine it. However, they still suffer from the following limitations: (a) Some methods use manually crafted predefined anchors, potentially affecting the model's generalization ability; (b) Most models primarily focus on encoding historical scene context, failing to fully utilize contextual cues from the future context where the predicted trajectories lie. By combining direct prediction with anchor-based methods, our future context encoding approach offers both flexibility and mitigation of mode collapse.

\subsection{Motion Forecasting via Trajectories}
Mainstream approaches in motion prediction~\cite{wang2023ganet,zhou2023query,gao2020vectornet,zhou2022hivt,varadarajan2022multipath++,shi2024mtr++,bhattacharyya2023ssl} model the future distribution of each agent by outputting a set of trajectories, which are represented as sequences of states (or state differences). An inherent characteristic of the driving environment is that vehicles and pedestrians can follow a variety of possible trajectories. To capture the uncertainty in agent motion, most methods predict multiple trajectories along with their confidence scores. Gaussian mixture models and Laplace mixture models are popular choices due to their compact parameterized forms. Some learning-based methods~\cite{wang2023ganet,liang2020learning} decode multiple trajectories directly from the encoded context, using training techniques or trajectory anchors to address the mode collapse problem. Anchor-based methods~\cite{chai2020multipath,gu2021densetnt} utilize predefined trajectory anchors or potential endpoints to model the discrete distribution. Predefined model methods~\cite{liu2021multimodal} allocate specific modules to predict different modes, while other approaches~\cite{gupta2018social,tang2019multiple} sample trajectory sets from a latent distribution and decode them using the model. HOME~\cite{naser2020heat} and GOHOME~\cite{gilles2022gohome} first predict a heatmap and then decode trajectories after sampling, whereas MP3~\cite{casas2021mp3} and NMP~\cite{zeng2019end} learn a cost function evaluator for trajectories, with the trajectories enumerated heuristically rather than generated by a learned model. Recently, some Transformer-based methods~\cite{wang2023ganet,zhou2023query,shi2024mtr++} have integrated anchor-free and anchor-based techniques to achieve state-of-the-art performance.

These trajectory-based output representation methods typically predict independent marginal distributions for each agent rather than the joint distribution of all agents coexisting in the spatial-temporal scene. Almost all work assumes an independent, per-agent output space, which cannot explicitly capture interactions between agents. A few studies describe joint interactions in the output, either asymmetrically~\cite{tolstaya2021identifying} or symmetrically~\cite{ngiam2021scene}. In the most challenging crowded scenarios, autonomous vehicles usually need to consider the future distributions of all targets simultaneously. These challenges make another form of motion output representation, occupancy grids, highly attractive.

\subsection{Motion Forecasting via Occupancy Grids.}
Currently, the use of occupancy grid representation for output in motion prediction is becoming increasingly popular~\cite{hu2023planning,mahjourian2022occupancy,casas2021mp3,hong2019rules,kim2022stopnet,agro2023implicit}. Occupancy grids predict the likelihood of discrete spatial-temporal grids being occupied in a bird's-eye view of the entire scene~\cite{mahjourian2022occupancy}. For example, ~\cite{hong2019rules} proposes a method to decode trajectories from occupancy grids, using a dynamic programming solution based on a simple motion model. MP3~\cite{casas2021mp3} introduces the concept of motion fields, predicting a set of forward motion vectors and the occupancy probability for each grid cell to aid in autonomous driving planning tasks. Similarly, Occupancy Flow Fields~\cite{mahjourian2022occupancy} extend the single occupancy grid probability by incorporating flow vector predictions, thereby introducing agent motion within the occupancy field. StopNet~\cite{kim2022stopnet} unifies trajectory prediction task and occupancy grid prediction task into a single model. Additionally, ~\cite{agro2023implicit} employs a single neural network to implicitly represent time-varying occupancy probabilities and flow vectors, allowing motion planners to query these directly at continuous spatial-temporal locations, thus avoiding unnecessary computations.

These occupancy grid methods primarily rely on two-dimensional rasterized image input representations, leading to heavy computational demands and limited compatibility with the currently dominant vector-based motion prediction methods. In these methods, scene information is typically rendered as a 2D bird's-eye view image, with different elements occupying different image channels. This allows for the direct use of mature CNN network models to encode scene context. However, these 2D image-based representations have inherent drawbacks: information loss during scene data rendering; limitations in the receptive field of convolutional networks, making them unsuitable for capturing complex dependencies in driving scenes, such as road topology extending along the road; and the high computational burden of grid-based modeling methods.

Our proposed Lane Occupancy Field method constructs lane occupancy fields based on vector representations derived from high-definition maps. This approach seamlessly integrates with existing vector-based trajectory prediction methods by adding a simple branch, requiring minimal additional computation when integrated with existing trajectory prediction frameworks. Moreover, compared to occupancy grids, lane occupancy fields provide lane-specific semantics, which are better suited to indicating drivable space in driving scenes and making full use of known scene map priors.
Since our proposed lane occupancy field can seamlessly integrate with vector-based trajectory prediction tasks, we propose a joint lane occupancy field prediction and trajectory prediction model. This model outputs two types of motion output representations, offering robust capabilities for capturing future motions.

\section{Results on the multi-world forecasting}
As shown in Table~\ref{test_multi_world}, we also submitted our test results on the Argoverse 2 multi-world forecasting dataset, and our method achieved state-of-the-art performance. In multi-world forecasting, we need to simultaneously predict a trajectory for each scored agent, forming a predicted world for multiple agents. In the Argoverse 2 multi-world forecasting dataset, the model is required to predict $K=6$ worlds for each scenario, along with the probability score for each world.

We report the official evaluation metrics, which include: the average minimum final displacement error (avgMinFDE) represents the mean FDE associated with a predicted world, summarized across all scored actors within a scenario. The world with the lowest avgMinFDE is referred to as the "best" world. 
The average minimum average displacement error (avgMinADE) denotes the mean ADE associated with a predicted world, summarized across all scored actors within a scenario.
The average brier minimum final displacement error (avgBrierMinFDE) signifies the mean brierMinFDE associated with a predicted world, summarized across all scored actors within a scenario.
The actor miss rate (actorMR) is defined over the evaluation set as the number of actor predictions considered to have "missed" (FDE > 2m) in the "best" (lowest minFDE) world, divided by the total number of scored actors.

Our method generates K trajectories for each predicted agent, resulting in $(C_6^1)^{N_A}$ possible combinations. For simplicity, we directly assign the K-th trajectory of each agent to the K-th predicted world, thereby aligning with the multi-world forecasting evaluation requirements. 

Despite not being specifically designed for multi-world forecasting, our model achieved competitive results on the multi-world forecasting dataset. Notably, our results on the official Argoverse 2 multi-world forecasting leaderboard significantly outperform FJMP~\cite{rowe2023fjmp}, a motion prediction method specifically designed for multi-agent interaction. Our model achieved improvements of 35.1\%, 33.9\%, 28.4\%, 41.5\%, and 36.8\% in avgBrierMinFDE6, avgMinFDE6, avgMinADE6, avgMinFDE1, and avgMinADE1, respectively.

\begin{table}[t]
\caption{Quantitative results on the Argoverse 2 multi-world motion forecasting leaderboard.}
\label{test_multi_world}
\begin{adjustbox}{width=\textwidth}
\centering
\begin{tabular}{c|cccc|cc}
\toprule
Method  & avgBrierMinFDE$_6$ $\downarrow$ & avgMinFDE$_6$ $\downarrow$	&avgMinADE$_6$ $\downarrow$ &actorMR$_6$ $\downarrow$	& avgMinFDE$_1$ $\downarrow$	&avgMinADE$_1$ $\downarrow$ \\
\midrule

FJMP~\cite{rowe2023fjmp} 	&2.59	&1.89	&0.81	&0.23	&4.00	&1.52    \\

Forecast-MAE~\cite{cheng2023forecast}	&2.24	&1.55	&0.69	&0.19	&3.33	&	1.30    \\

HeteroGCN++~\cite{gao2023dynamic}	&2.12	&1.46	&0.69	&0.19	&3.05	&1.23  \\

\textbf{FutureNet-LOF (Ours)}	& \textbf{1.68}	& \textbf{1.25}	& \textbf{0.58}	& \textbf{0.18}	&\textbf{2.34}	&\textbf{0.96}    \\
\bottomrule
\end{tabular}
\end{adjustbox}
\end{table}

\section{More ablation results}
We evaluate the prediction performance of our FutureNet model across different prediction horizons. As shown in Table~\ref{ablation_horizen}, the first row represents the baseline model, which predicts 6-second trajectories in a single step using trajectory queries encoded with scene information during decoding. In this baseline model, we do not use the recurrent prediction and refinement prediction decoder with future scene context encoding. In the second row, we utilize a recurrent decoder with a recurrence step of 3, where the trajectory query is re-encoded based on future context information during each recurrent prediction step. The third row represents a FutureNet model, which employs both recurrent prediction and refinement prediction modules with future context encoding.

We calculate the minFDE6 and MR6 metrics at 2s, 4s, and 6s for the prediction results of the three models. At the prediction horizon of 2s, the performance of the three models is similar. This aligns with our intuition, as our FutureNet model treats 2s as a keyframe and re-encodes the future scene after the initial prediction of the first 2s of the trajectory. At 4s, our model with the future context encoding module improves the minFDE6 and MR6 performance of the baseline model by 5.7\% and 9.6\%, respectively. At 6s, our model with the future encoding module further significantly enhances performance, with improvements of 12.5\% and 23.4\%. The experimental results demonstrate that our FutureNet with the future encoding module exhibits superior performance in long-term prediction.

\begin{table}[ht]\small
\caption{Ablation studies on different prediction horizons, based on the Argoverse 2 validation set.}
\label{ablation_horizen}
\begin{adjustbox}{width=\textwidth}
\centering
\begin{tabular}{ccc|c|ccc}
\toprule
{One step} & {Recurrence} & {Refinement} & {horizon} &\makecell[c]{b-minFDE$_6$$\downarrow$} &\makecell[c]{minFDE$_6$$\downarrow$} & \makecell[c]{MR$_6$$\downarrow$}   \\
\midrule
\multirow{3}{*}{\checkmark} & \multirow{3}{*}{} & \multirow{3}{*}{} & 2s & 0.92 & \textbf{0.27} & 0.006 \\
&&&  4s & 1.30 & 0.70& 0.052 \\ 
&&&  6s & 1.95 & 1.36 & 0.184\\ 
\midrule
\multirow{3}{*}{} & \multirow{3}{*}{\checkmark}& \multirow{3}{*}{}  & 2s& 0.93 & \textbf{0.27} & 0.006  \\
&&&  4s& 1.28 & \textbf{0.66} & \textbf{0.047} \\ 
&&&  6s& 1.82 & 1.21 & 0.148 \\ 
\midrule
\multirow{3}{*}{} &  \multirow{3}{*}{\checkmark} & \multirow{3}{*}{\checkmark}  & 2s &\textbf{0.91}& \textbf{0.27}  & \textbf{0.005} \\
&&&  4s &\textbf{1.27}& \textbf{0.66}  &  \textbf{0.047} \\ 
&&&  6s &\textbf{1.80}& \textbf{1.19}  & \textbf{0.141}  \\ 
\bottomrule
\end{tabular}
\end{adjustbox}
\end{table}

\section{Comparison with state-of-the-art methods}
\begin{table}[t]
\caption{Quantitative results on the Argoverse 1 motion forecasting leaderboard.}
\label{test1}
\begin{adjustbox}{width=\textwidth}
\centering
\begin{tabular}{c|cccc|ccc}
\toprule
Method    & b-minFDE$_6$ $\downarrow$	 &minFDE$_6$ $\downarrow$	&minADE$_6$ $\downarrow$	&MR$_6$ $\downarrow$	&minFDE$_1$ $\downarrow$	& minADE$_1$ $\downarrow$	&MR$_1$ $\downarrow$ \\

\midrule
DenseTNT~\cite{gu2021densetnt}	&1.98	&1.28	&0.88	&0.13	&3.63 &	1.68	&0.58 \\
HiVT++~\cite{zhou2022hivt}	&1.82	&1.15	&0.77	&0.12	&3.44	&1.56	&0.54  \\
TPCN++~\cite{ye2021tpcn}	&1.80	&1.17	&0.78	&0.12	&3.33	&1.51	&0.54  \\
multipath++~\cite{varadarajan2022multipath++}	&1.79	&1.21	&0.79	&0.13	&3.61	&1.62	&0.56  \\
GANet~\cite{wang2023ganet} 	&1.79	&1.16	&0.81	&0.12	&3.45	&1.59	&0.55  \\
macformer~\cite{feng2023macformer} 	&1.77 	&1.21	&0.81	&0.13	&3.61	&1.66	&0.56  \\
DCMS~\cite{ye2023bootstrap}	&1.76	&1.14	&0.77	&0.11	&\textbf{3.25}	&\textbf{1.48}	&0.53  \\
HeteroGCN~\cite{gao2023dynamic} 	&1.75	&1.16	&0.79	&0.12	&3.41	&1.57	&0.54  \\
Wayformer~\cite{nayakanti2023wayformer} 	&1.74 	&1.16	&0.77	&0.12	&3.66	&1.64	&0.57  \\
ProphNet~\cite{wang2023prophnet} 	&1.69	&1.13	&0.76	&0.11	&3.26	&1.49	&0.53  \\
QCNet~\cite{zhou2023query} 	&1.69	&1.07	&\textbf{0.73}	&0.11	&3.34	&1.52	&0.53  \\

\textbf{FutureNet-LOF (Ours)}	 	& \textbf{1.66}	&\textbf{1.03}		&\textbf{0.73}	&\textbf{0.10} 	&\textbf{3.25}		&1.50	&\textbf{0.51}		\\	
\bottomrule
\end{tabular}
\end{adjustbox}
\end{table}

QCNet is a state-of-the-art method on two large-scale autonomous driving motion prediction datasets, Argoverse 1 and Argoverse 2. We equip the QCNet method with our proposed FutureNet-LOF and conduct a comprehensive comparison with QCNet under the same training and evaluation settings on the Argoverse 2 validation dataset. As shown in Table~\ref{qcnet_comp}, our method outperforms QCNet across all evaluation metrics. Specifically, QCNet's performance is 16.3\% and 10.4\% worse than ours on the MR6 and MR1 metrics, respectively, indicating our method's superior ability to capture the diverse future movements of agents. Additionally, QCNet scores 7.0\% and 8.0\% lower than our method on the minFDE6 and minFDE1 metrics, respectively. This demonstrates that our method achieves higher predictive accuracy, with predicted trajectories more closely aligning with the ground truth.
\begin{table}[H]
\caption{Comparison of results between FutureNet-LOF and QCNet on the Argoverse 2 validation dataset.}
\label{qcnet_comp}
\begin{adjustbox}{width=\textwidth}
\centering
\begin{tabular}{c|cccc|ccc}
\toprule
Method    & b-minFDE$_6$ $\downarrow$	 &minFDE$_6$ $\downarrow$	&minADE$_6$ $\downarrow$	&MR$_6$ $\downarrow$	&minFDE$_1$ $\downarrow$	& minADE$_1$ $\downarrow$	&MR$_1$ $\downarrow$ \\
\midrule
QCNet~\cite{zhou2023query}	&1.860	&1.243	&0.718	&0.157	&4.285	&1.676	&0.603  \\
\textbf{FutureNet-LOF (Ours)}	& \textbf{1.792}	& \textbf{1.162}	& \textbf{0.705}	& \textbf{0.135}	& \textbf{3.968}	&\textbf{1.590}	&\textbf{0.546}    \\
\bottomrule
\end{tabular}
\end{adjustbox}
\end{table}

MTR++~\cite{shi2024mtr++} is a state-of-the-art method on the large-scale Waymo Open Dataset~\cite{sun2020scalability}. We report our submitted results' performance metrics on the official Argoverse 2 motion forecasting leaderboard in Table~\ref{mtr++_comp}. The results show that MTR++ underperforms our method by more than 12\%, 25\%, 20\%, 13\%, and 11\% in terms of b-minFDE6, minFDE6, minADE6, minFDE1, and minADE1, respectively.

\begin{table}[H]
\caption{Performance comparison of MTR++~\cite{shi2024mtr++} and FutureNet-LOF methods on the Argoverse 2 test dataset.}
\label{mtr++_comp}
\begin{adjustbox}{width=\textwidth}
\centering
\begin{tabular}{c|cccc|ccc}
\toprule
Method    & b-minFDE$_6$ $\downarrow$	 &minFDE$_6$ $\downarrow$	&minADE$_6$ $\downarrow$	&MR$_6$ $\downarrow$	&minFDE$_1$ $\downarrow$	& minADE$_1$ $\downarrow$	&MR$_1$ $\downarrow$ \\
\midrule
MTR++~\cite{shi2024mtr++}	&1.88	&1.37	&0.71	&0.14	&4.12	&1.64	&0.56  \\
\textbf{FutureNet-LOF (Ours)}	& \textbf{1.63}	& \textbf{1.07}	& \textbf{0.58}	& \textbf{0.12}	& \textbf{3.63}	&\textbf{1.46}	&\textbf{0.51}    \\
\bottomrule
\end{tabular}
\end{adjustbox}
\end{table}

Since the approach of our multiple parallel local worlds that we propose is similar to the query-centric modeling in QCNet~\cite{zhou2023query} and MTR++~\cite{shi2024mtr++}, we provide a simple comparison of these two modeling methods in Table~\ref{mtr++_query}. Conceptually, query-centric modeling can be seen as a specific instance of our more general multiple local world encoding approach.

Our multiple parallel local worlds modeling approach is more intuitive than the method used in MTR++. We can observe our own movement patterns while walking, as each of us navigates within a global scene from a self-centered perspective. In this framework, each individual constitutes a local world, and each scene element is also considered a local world, anchored in the global scene by coordinates and heading. Each world may possess a hierarchical structure, with atomic-level local worlds forming higher-level worlds. For instance, trajectory state worlds at each time step form a trajectory-level local world, and map sampling point worlds form a map polygon-level world. The various possible future trajectories also represent parallel future local worlds.

Compared to current query-centric modeling methods such as MTR++~\cite{shi2024mtr++} and QCNet~\cite{zhou2023query}, our approach is more comprehensive. We consider every element in the scene as a local world, encompassing various levels of granularity. For example, our method includes elements at the map polygon level as well as the map sampled point level; it considers agent trajectories as well as the state of each trajectory at every timestep. Additionally, it incorporates multiple parallel trajectory query elements and map query elements.

Compared to MTR++'s query-centric modeling~\cite{shi2024mtr++}, our approach is more intuitive and adaptive.
(a) Manual preprocessing: MTR++ uses learnable intention queries to identify an agent's potential motion intentions by generating K representative intention points through manual preprocessing with the k-means clustering algorithm on the endpoints of ground-truth trajectories in the training dataset. This process not only requires significant manual preprocessing but also hinders the model's ability to effectively adapt based on scene information. In contrast, our approach performs direct recurrent prediction without relying on manual preprocessing, resulting in a more flexible and adaptive model. We encode the future scenario based on the preliminary predicted trajectories, placing the predicted trajectories into the future scenario.
(b) Query design: MTR++'s intention query has some shortcomings. When conducting query-centric interaction modeling, MTR++ transforms all scene information into the coordinate system of the query feature. However, because their intention query is derived from clustering on the dataset to obtain preprocessed coordinates, this transformation is not feasible. Consequently, they assign the heading of the current state of the target trajectory to all K queries, resulting in a suboptimal design. Our approach, on the other hand, predicts future trajectories in a recurrent manner and uses the endpoint coordinates and heading of the trajectories as new anchors in the global world, making it more flexible and intuitive.

\begin{table}[htb]
\caption{Multiple parallel local worlds of our approach vs. MTR++ ‘s query centric modeling~\cite{shi2024mtr++}
}
\label{mtr++_query}
\begin{adjustbox}{width=\textwidth}
\centering
\begin{tabular}{p{3cm}|p{6cm}|p{6cm}}
\toprule
&Multiple parallel local worlds modeling & Query centric modeling \\
\midrule
&\multicolumn{2}{c}{Scene elements Modeling} \\
\midrule
Elements & Map point elements, 
\newline Map polygon elements, 
\newline Multiple trajectory level elements at each time step,
\newline Multiple proposal trajectory query elements at each recurrent step,
\newline Refinement trajectory query elements,
\newline Map query elements.
& Map polygon elements,
\newline Trajectory level elements,
\newline Intent query elements. \\
\midrule
&\multicolumn{2}{c}{Query Modeling} \\
\midrule
 The purpose of query modeling & Encode the future context based on preliminary prediction and make prediction.
& Pinpoint an agent’s potential motion intentions and make prediction. \\
\midrule
Global anchor for trajectory query &
Global anchor of the query is the coordinate and heading of the known (observed or predicted ) trajectory’s latest state.
& At the beginning, it utilizes manually predefined coordinates, assigning the current heading of the target trajectory to K queries as their heading. \\
\midrule
Core design &
1. Directly predict the future trajectory and decode the trajectory endpoint's coordinates and heading as the trajectory query's anchor.
2. Encode the future context for further prediction using the anchor, essentially placing the predicted anchor into the future context for new query encoding.
Trajectory query local worlds are adaptively constructed based on scene information during the recurrent and refinement processes..
& 1. Using a manual k-means algorithm to cluster the real trajectory endpoints in the training dataset to obtain initial coordinates. \newline
 2. Assigning the current heading of the target trajectory state to all K queries as their heading direction. \\
 \midrule
 Multiple query & Each recurrent prediction or refinement prediction has a specific query.
& No recurrent multiple query. \\
\midrule
Map query & Use map query for lane occupancy prediction
& No map query \\
\midrule
Conclusion &  Intuitive, comprehensive, and adaptive. & Deficient and dependent on the manually crafted preprocessing \\
\bottomrule
\end{tabular}
\end{adjustbox}
\end{table}

\section{Training implementation and details}
We conduct parallel training on 8 A100 GPUs using the AdamW optimizer.
The batch size is set to 32, with an initial learning rate of $5\times e^{-4}$, and a weight decay of $1 \times 10^{-4}$. The learning rate decays using the cosine annealing scheduler~\cite{loshchilov2016sgdr}. We adopt 3 recurrent steps for the FutureNet-LOF model. The map encoding attention layer consists of 1 layer. For the agent encoder and trajectory decoder, we use 2 attention layers for most attention modules, while the mode-to-mode attention consists of 1 layer. All multi-head attention blocks have 8 heads. The hidden dimension of our model is 128. In the loss function, we empirically set $\beta = 1$ and $\rho = 20$.

The radius for map-to-map polygon interaction is 150 meters. For the temporal attention mechanism, the time step span is set to 10. The neighbor range for map-to-agent interaction is 50 meters, and the neighbor range for agent-to-agent interaction is also 50 meters. For the temporal-to-trajectory query attention mechanism, the time step span is 30. The neighbor range for map-to-trajectory query interaction and agent-to-trajectory query interaction is 150 meters each. These radii of local neighbor regions are similar to those in QCNet~\cite{zhou2023query}. The range for trajectory query-to-map query interaction and map query-to-trajectory query interaction is set to 10 meters. For the results on test set, we also use an ensemble method similar to that in QCNet~\cite{zhou2023query}.

Our FutureNet-LOF model's inference speed is 50 scenarios per second on a single A100 GPU.

\include{detailed_net}

\section{Limitations}
Like all other prediction methods, our method also faces the challenge of the long tail problem. For extremely abnormal driving cases, the predicted results may fail to capture the actor's actual future trajectory, e.g., an actor keeps going straight in the observed history but suddenly switches lanes in the future. The randomness of the driver's driving intention brings significant challenges to all prediction methods, and further improvements are needed to alleviate the long tail problem.

\section{Broader impacts}
\subsection{Potential positive societal impacts}
Autonomous driving technology has the potential to revolutionize the transportation industry, offering significant value in various high-potential markets such as transportation, freight logistics, and autonomous robots. Research and reports indicate that autonomous driving technology can greatly enhance road safety and reduce traffic accidents. In intelligent driving algorithms, motion prediction serves as a crucial link between perception and planning, playing an essential role in ensuring the safety of the entire autonomous driving system. Autonomous vehicles rely on accurate predictions of other road users' future behaviors to make safe and rational decisions in subsequent planning and control modules. Therefore, precise motion prediction is vital for safe and smooth autonomous driving. Our method aims to improve the accuracy of motion prediction in autonomous driving, which is critical for achieving highly reliable and safe autonomous driving.
\subsection{Potential negative societal impacts}
The potential negative societal impacts may come from the aforementioned long tail problem that all motion forecasting methods face. 
A self-driving car may make an unreasonable decision in the subsequent planning and control module when a big prediction failure occurs, where a possibility has not been considered, and all modalities miss the actual future.
Fortunately, motion prediction algorithms are performed with high frequency in real applications so that those failed predictions can be corrected rapidly and continuously.
Although our method has improved a lot on these challenging cases compared to previous works, the long tail problem needs more studies and explorations.


\newpage
\section*{NeurIPS Paper Checklist}

\begin{enumerate}

\item {\bf Claims}
    \item[] Question: Do the main claims made in the abstract and introduction accurately reflect the paper's contributions and scope?
    \item[] Answer: \answerYes{}
    \item[] Justification: We propose FutureNet-LOF, which has been discussed in the abstract and introduction.
    \item[] Guidelines: 
    \begin{itemize}
        \item The answer NA means that the abstract and introduction do not include the claims made in the paper.
        \item The abstract and/or introduction should clearly state the claims made, including the contributions made in the paper and important assumptions and limitations. A No or NA answer to this question will not be perceived well by the reviewers. 
        \item The claims made should match theoretical and experimental results, and reflect how much the results can be expected to generalize to other settings. 
        \item It is fine to include aspirational goals as motivation as long as it is clear that these goals are not attained by the paper. 
    \end{itemize}

\item {\bf Limitations}
    \item[] Question: Does the paper discuss the limitations of the work performed by the authors?
    \item[] Answer: \answerYes{} 
    \item[] Justification: Refer limitations section.
    \item[] Guidelines:
    \begin{itemize}
        \item The answer NA means that the paper has no limitation while the answer No means that the paper has limitations, but those are not discussed in the paper. 
        \item The authors are encouraged to create a separate "Limitations" section in their paper.
        \item The paper should point out any strong assumptions and how robust the results are to violations of these assumptions (e.g., independence assumptions, noiseless settings, model well-specification, asymptotic approximations only holding locally). The authors should reflect on how these assumptions might be violated in practice and what the implications would be.
        \item The authors should reflect on the scope of the claims made, e.g., if the approach was only tested on a few datasets or with a few runs. In general, empirical results often depend on implicit assumptions, which should be articulated.
        \item The authors should reflect on the factors that influence the performance of the approach. For example, a facial recognition algorithm may perform poorly when image resolution is low or images are taken in low lighting. Or a speech-to-text system might not be used reliably to provide closed captions for online lectures because it fails to handle technical jargon.
        \item The authors should discuss the computational efficiency of the proposed algorithms and how they scale with dataset size.
        \item If applicable, the authors should discuss possible limitations of their approach to address problems of privacy and fairness.
        \item While the authors might fear that complete honesty about limitations might be used by reviewers as grounds for rejection, a worse outcome might be that reviewers discover limitations that aren't acknowledged in the paper. The authors should use their best judgment and recognize that individual actions in favor of transparency play an important role in developing norms that preserve the integrity of the community. Reviewers will be specifically instructed to not penalize honesty concerning limitations.
    \end{itemize}

\item {\bf Theory Assumptions and Proofs}
    \item[] Question: For each theoretical result, does the paper provide the full set of assumptions and a complete (and correct) proof?
    \item[] Answer: \answerYes{} 
    \item[] Justification: 
    \item[] Guidelines:
    \begin{itemize}
        \item The answer NA means that the paper does not include theoretical results. 
        \item All the theorems, formulas, and proofs in the paper should be numbered and cross-referenced.
        \item All assumptions should be clearly stated or referenced in the statement of any theorems.
        \item The proofs can either appear in the main paper or the supplemental material, but if they appear in the supplemental material, the authors are encouraged to provide a short proof sketch to provide intuition. 
        \item Inversely, any informal proof provided in the core of the paper should be complemented by formal proofs provided in appendix or supplemental material.
        \item Theorems and Lemmas that the proof relies upon should be properly referenced. 
    \end{itemize}

    \item {\bf Experimental Result Reproducibility}
    \item[] Question: Does the paper fully disclose all the information needed to reproduce the main experimental results of the paper to the extent that it affects the main claims and/or conclusions of the paper (regardless of whether the code and data are provided or not)?
    \item[] Answer: \answerYes{} 
    \item[] Justification: Refer approach and training implementation and details section.
    \item[] Guidelines:
    \begin{itemize}
        \item The answer NA means that the paper does not include experiments.
        \item If the paper includes experiments, a No answer to this question will not be perceived well by the reviewers: Making the paper reproducible is important, regardless of whether the code and data are provided or not.
        \item If the contribution is a dataset and/or model, the authors should describe the steps taken to make their results reproducible or verifiable. 
        \item Depending on the contribution, reproducibility can be accomplished in various ways. For example, if the contribution is a novel architecture, describing the architecture fully might suffice, or if the contribution is a specific model and empirical evaluation, it may be necessary to either make it possible for others to replicate the model with the same dataset, or provide access to the model. In general. releasing code and data is often one good way to accomplish this, but reproducibility can also be provided via detailed instructions for how to replicate the results, access to a hosted model (e.g., in the case of a large language model), releasing of a model checkpoint, or other means that are appropriate to the research performed.
        \item While NeurIPS does not require releasing code, the conference does require all submissions to provide some reasonable avenue for reproducibility, which may depend on the nature of the contribution. For example
        \begin{enumerate}
            \item If the contribution is primarily a new algorithm, the paper should make it clear how to reproduce that algorithm.
            \item If the contribution is primarily a new model architecture, the paper should describe the architecture clearly and fully.
            \item If the contribution is a new model (e.g., a large language model), then there should either be a way to access this model for reproducing the results or a way to reproduce the model (e.g., with an open-source dataset or instructions for how to construct the dataset).
            \item We recognize that reproducibility may be tricky in some cases, in which case authors are welcome to describe the particular way they provide for reproducibility. In the case of closed-source models, it may be that access to the model is limited in some way (e.g., to registered users), but it should be possible for other researchers to have some path to reproducing or verifying the results.
        \end{enumerate}
    \end{itemize}

\item {\bf Open access to data and code}
    \item[] Question: Does the paper provide open access to the data and code, with sufficient instructions to faithfully reproduce the main experimental results, as described in supplemental material?
    \item[] Answer: \answerYes{} 
    \item[] Justification: 
    \item[] Guidelines:
    \begin{itemize}
        \item The answer NA means that paper does not include experiments requiring code.
        \item Please see the NeurIPS code and data submission guidelines (\url{https://nips.cc/public/guides/CodeSubmissionPolicy}) for more details.
        \item While we encourage the release of code and data, we understand that this might not be possible, so “No” is an acceptable answer. Papers cannot be rejected simply for not including code, unless this is central to the contribution (e.g., for a new open-source benchmark).
        \item The instructions should contain the exact command and environment needed to run to reproduce the results. See the NeurIPS code and data submission guidelines (\url{https://nips.cc/public/guides/CodeSubmissionPolicy}) for more details.
        \item The authors should provide instructions on data access and preparation, including how to access the raw data, preprocessed data, intermediate data, and generated data, etc.
        \item The authors should provide scripts to reproduce all experimental results for the new proposed method and baselines. If only a subset of experiments are reproducible, they should state which ones are omitted from the script and why.
        \item At submission time, to preserve anonymity, the authors should release anonymized versions (if applicable).
        \item Providing as much information as possible in supplemental material (appended to the paper) is recommended, but including URLs to data and code is permitted.
    \end{itemize}

\item {\bf Experimental Setting/Details}
    \item[] Question: Does the paper specify all the training and test details (e.g., data splits, hyperparameters, how they were chosen, type of optimizer, etc.) necessary to understand the results?
    \item[] Answer: \answerYes{} 
    \item[] Justification: Refer approach and training implementation and details section.
    \item[] Guidelines:
    \begin{itemize}
        \item The answer NA means that the paper does not include experiments.
        \item The experimental setting should be presented in the core of the paper to a level of detail that is necessary to appreciate the results and make sense of them.
        \item The full details can be provided either with the code, in appendix, or as supplemental material.
    \end{itemize}

\item {\bf Experiment Statistical Significance}
    \item[] Question: Does the paper report error bars suitably and correctly defined or other appropriate information about the statistical significance of the experiments?
    \item[] Answer: \answerYes{} 
    \item[] Justification: Refer experiment section.
    \item[] Guidelines:
    \begin{itemize}
        \item The answer NA means that the paper does not include experiments.
        \item The authors should answer "Yes" if the results are accompanied by error bars, confidence intervals, or statistical significance tests, at least for the experiments that support the main claims of the paper.
        \item The factors of variability that the error bars are capturing should be clearly stated (for example, train/test split, initialization, random drawing of some parameter, or overall run with given experimental conditions).
        \item The method for calculating the error bars should be explained (closed form formula, call to a library function, bootstrap, etc.)
        \item The assumptions made should be given (e.g., Normally distributed errors).
        \item It should be clear whether the error bar is the standard deviation or the standard error of the mean.
        \item It is OK to report 1-sigma error bars, but one should state it. The authors should preferably report a 2-sigma error bar than state that they have a 96\% CI, if the hypothesis of Normality of errors is not verified.
        \item For asymmetric distributions, the authors should be careful not to show in tables or figures symmetric error bars that would yield results that are out of range (e.g. negative error rates).
        \item If error bars are reported in tables or plots, The authors should explain in the text how they were calculated and reference the corresponding figures or tables in the text.
    \end{itemize}

\item {\bf Experiments Compute Resources}
    \item[] Question: For each experiment, does the paper provide sufficient information on the computer resources (type of compute workers, memory, time of execution) needed to reproduce the experiments?
    \item[] Answer: \answerYes{} 
    \item[] Justification: Refer training implementation and details section.
    \item[] Guidelines:
    \begin{itemize}
        \item The answer NA means that the paper does not include experiments.
        \item The paper should indicate the type of compute workers CPU or GPU, internal cluster, or cloud provider, including relevant memory and storage.
        \item The paper should provide the amount of compute required for each of the individual experimental runs as well as estimate the total compute. 
        \item The paper should disclose whether the full research project required more compute than the experiments reported in the paper (e.g., preliminary or failed experiments that didn't make it into the paper). 
    \end{itemize}
    
\item {\bf Code Of Ethics}
    \item[] Question: Does the research conducted in the paper conform, in every respect, with the NeurIPS Code of Ethics \url{https://neurips.cc/public/EthicsGuidelines}?
    \item[] Answer: \answerYes{} 
    \item[] Justification: 
    \item[] Guidelines:
    \begin{itemize}
        \item The answer NA means that the authors have not reviewed the NeurIPS Code of Ethics.
        \item If the authors answer No, they should explain the special circumstances that require a deviation from the Code of Ethics.
        \item The authors should make sure to preserve anonymity (e.g., if there is a special consideration due to laws or regulations in their jurisdiction).
    \end{itemize}

\item {\bf Broader Impacts}
    \item[] Question: Does the paper discuss both potential positive societal impacts and negative societal impacts of the work performed?
    \item[] Answer: \answerYes{} 
    \item[] Justification: Refer broader impacts secton.
    \item[] Guidelines:
    \begin{itemize}
        \item The answer NA means that there is no societal impact of the work performed.
        \item If the authors answer NA or No, they should explain why their work has no societal impact or why the paper does not address societal impact.
        \item Examples of negative societal impacts include potential malicious or unintended uses (e.g., disinformation, generating fake profiles, surveillance), fairness considerations (e.g., deployment of technologies that could make decisions that unfairly impact specific groups), privacy considerations, and security considerations.
        \item The conference expects that many papers will be foundational research and not tied to particular applications, let alone deployments. However, if there is a direct path to any negative applications, the authors should point it out. For example, it is legitimate to point out that an improvement in the quality of generative models could be used to generate deepfakes for disinformation. On the other hand, it is not needed to point out that a generic algorithm for optimizing neural networks could enable people to train models that generate Deepfakes faster.
        \item The authors should consider possible harms that could arise when the technology is being used as intended and functioning correctly, harms that could arise when the technology is being used as intended but gives incorrect results, and harms following from (intentional or unintentional) misuse of the technology.
        \item If there are negative societal impacts, the authors could also discuss possible mitigation strategies (e.g., gated release of models, providing defenses in addition to attacks, mechanisms for monitoring misuse, mechanisms to monitor how a system learns from feedback over time, improving the efficiency and accessibility of ML).
    \end{itemize}
    
\item {\bf Safeguards}
    \item[] Question: Does the paper describe safeguards that have been put in place for responsible release of data or models that have a high risk for misuse (e.g., pretrained language models, image generators, or scraped datasets)?
    \item[] Answer: \answerNA{} 
    \item[] Justification: 
    \item[] Guidelines:
    \begin{itemize}
        \item The answer NA means that the paper poses no such risks.
        \item Released models that have a high risk for misuse or dual-use should be released with necessary safeguards to allow for controlled use of the model, for example by requiring that users adhere to usage guidelines or restrictions to access the model or implementing safety filters. 
        \item Datasets that have been scraped from the Internet could pose safety risks. The authors should describe how they avoided releasing unsafe images.
        \item We recognize that providing effective safeguards is challenging, and many papers do not require this, but we encourage authors to take this into account and make a best faith effort.
    \end{itemize}

\item {\bf Licenses for existing assets}
    \item[] Question: Are the creators or original owners of assets (e.g., code, data, models), used in the paper, properly credited and are the license and terms of use explicitly mentioned and properly respected?
    \item[] Answer: \answerYes{} 
    \item[] Justification: 
    \item[] Guidelines:
    \begin{itemize}
        \item The answer NA means that the paper does not use existing assets.
        \item The authors should cite the original paper that produced the code package or dataset.
        \item The authors should state which version of the asset is used and, if possible, include a URL.
        \item The name of the license (e.g., CC-BY 4.0) should be included for each asset.
        \item For scraped data from a particular source (e.g., website), the copyright and terms of service of that source should be provided.
        \item If assets are released, the license, copyright information, and terms of use in the package should be provided. For popular datasets, \url{paperswithcode.com/datasets} has curated licenses for some datasets. Their licensing guide can help determine the license of a dataset.
        \item For existing datasets that are re-packaged, both the original license and the license of the derived asset (if it has changed) should be provided.
        \item If this information is not available online, the authors are encouraged to reach out to the asset's creators.
    \end{itemize}

\item {\bf New Assets}
    \item[] Question: Are new assets introduced in the paper well documented and is the documentation provided alongside the assets?
    \item[] Answer: \answerNA{} 
    \item[] Justification: 
    \item[] Guidelines:
    \begin{itemize}
        \item The answer NA means that the paper does not release new assets.
        \item Researchers should communicate the details of the dataset/code/model as part of their submissions via structured templates. This includes details about training, license, limitations, etc. 
        \item The paper should discuss whether and how consent was obtained from people whose asset is used.
        \item At submission time, remember to anonymize your assets (if applicable). You can either create an anonymized URL or include an anonymized zip file.
    \end{itemize}

\item {\bf Crowdsourcing and Research with Human Subjects}
    \item[] Question: For crowdsourcing experiments and research with human subjects, does the paper include the full text of instructions given to participants and screenshots, if applicable, as well as details about compensation (if any)? 
    \item[] Answer: \answerNA{} 
    \item[] Justification: 
    \item[] Guidelines:
    \begin{itemize}
        \item The answer NA means that the paper does not involve crowdsourcing nor research with human subjects.
        \item Including this information in the supplemental material is fine, but if the main contribution of the paper involves human subjects, then as much detail as possible should be included in the main paper. 
        \item According to the NeurIPS Code of Ethics, workers involved in data collection, curation, or other labor should be paid at least the minimum wage in the country of the data collector. 
    \end{itemize}

\item {\bf Institutional Review Board (IRB) Approvals or Equivalent for Research with Human Subjects}
    \item[] Question: Does the paper describe potential risks incurred by study participants, whether such risks were disclosed to the subjects, and whether Institutional Review Board (IRB) approvals (or an equivalent approval/review based on the requirements of your country or institution) were obtained?
    \item[] Answer: \answerNA{} 
    \item[] Justification: 
    \item[] Guidelines:
    \begin{itemize}
        \item The answer NA means that the paper does not involve crowdsourcing nor research with human subjects.
        \item Depending on the country in which research is conducted, IRB approval (or equivalent) may be required for any human subjects research. If you obtained IRB approval, you should clearly state this in the paper. 
        \item We recognize that the procedures for this may vary significantly between institutions and locations, and we expect authors to adhere to the NeurIPS Code of Ethics and the guidelines for their institution. 
        \item For initial submissions, do not include any information that would break anonymity (if applicable), such as the institution conducting the review.
    \end{itemize}

\end{enumerate}

\end{document}

%% file: detailed_net.tex
\section{Detailed diagrams of unfolded FutureNet-LOF network pipeline}
\label{unfolded}
\subsection{Multiple parallel local worlds modeling} 
\label{multi_local_worlds}
We create a local world for each scene element, incorporating global coordinate and heading.
\begin{figure}[H]
    \centering
    \begin{minipage}{0.32\textwidth}
\captionsetup{labelformat=empty}
        \caption{Scene context.}
        \vspace{-6px}
        \centering
        \includegraphics[width=0.8\textwidth]{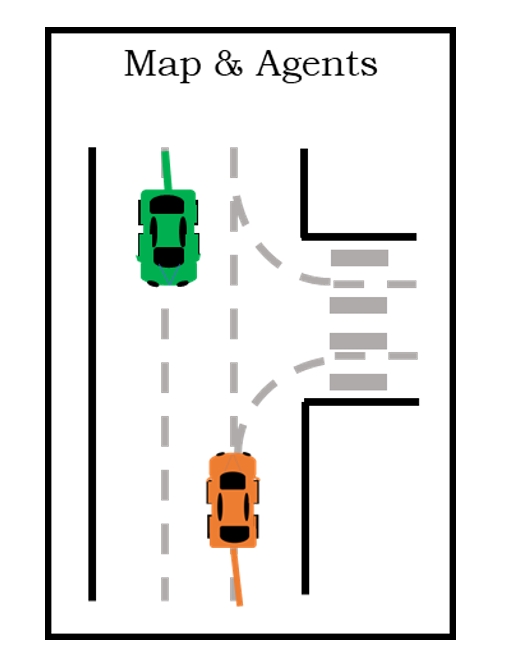}
    \end{minipage}
    \begin{minipage}{0.32\textwidth}
\captionsetup{labelformat=empty,justification=raggedright,singlelinecheck=false}
        \caption{(1) Map point elements.}
        \centering
        \includegraphics[width=0.8\textwidth]{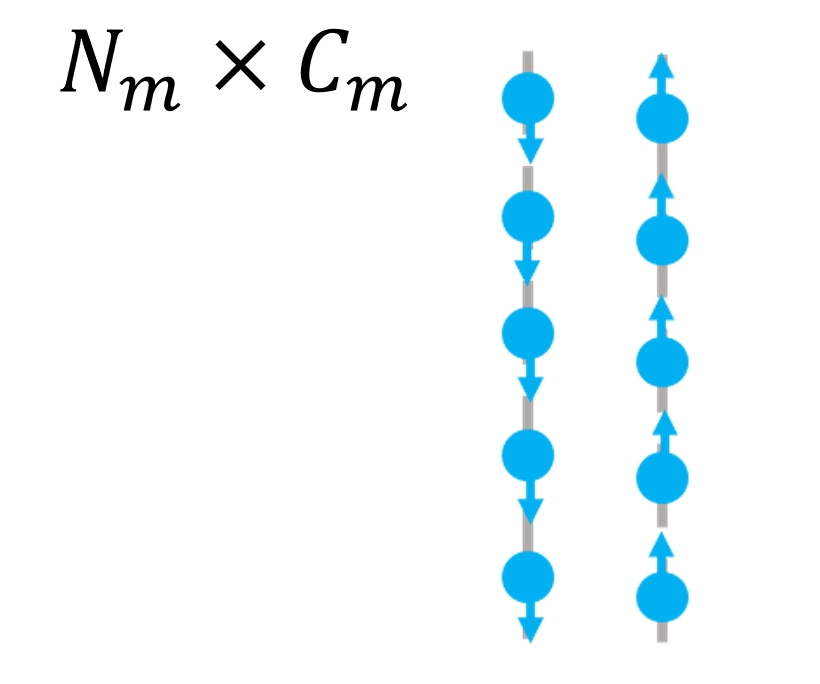}
    \end{minipage}
    \begin{minipage}{0.32\textwidth}
        \captionsetup{labelformat=empty}
        \caption{(2) Map polygon elements.}
        \centering
        \includegraphics[width=0.8\textwidth]{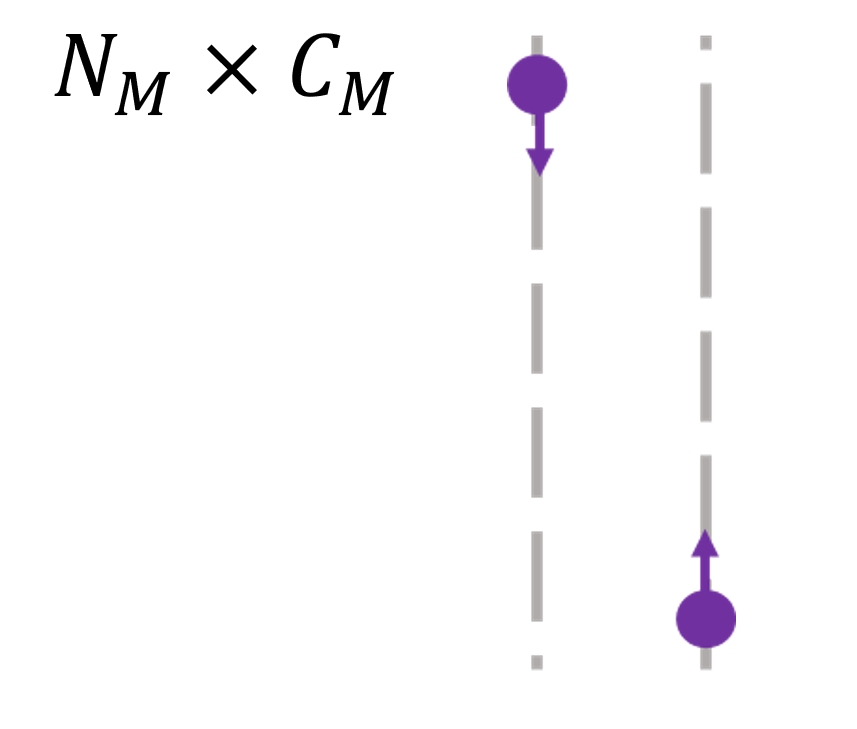}
    \end{minipage}
\end{figure}

\begin{figure}[H]
\captionsetup{labelformat=empty,justification=raggedright,singlelinecheck=false}
\caption{(3) Multiple trajectory level elements at each time step.}
\centering
\includegraphics[width=0.9\textwidth]{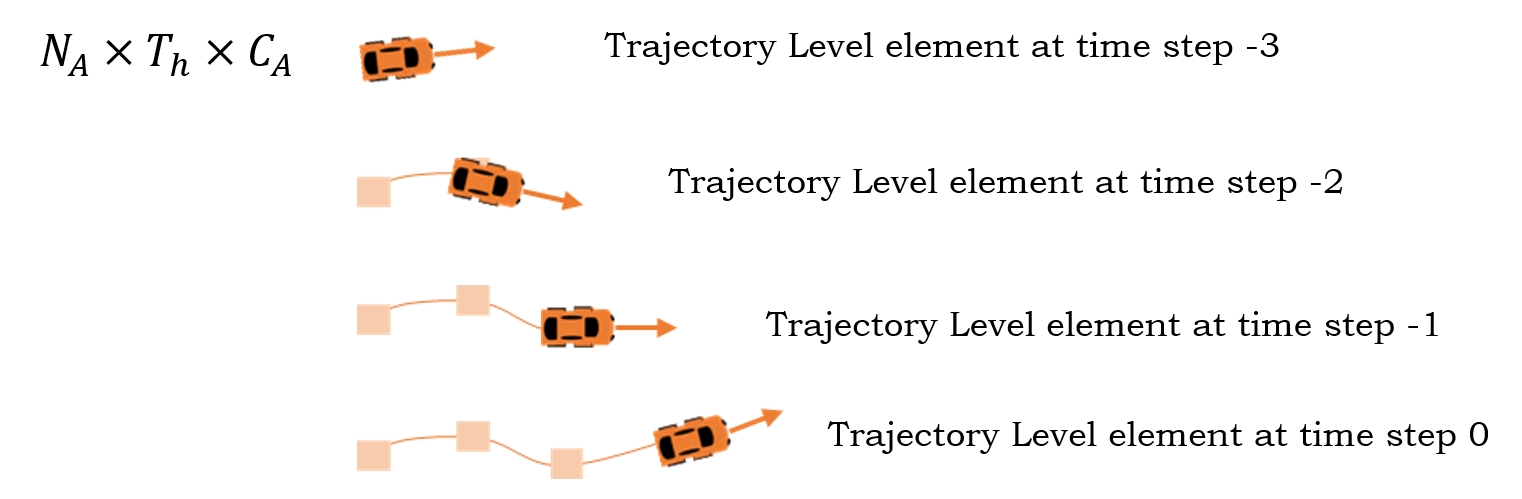}
\end{figure}

\begin{figure}[H]
\centering
\captionsetup{labelformat=empty,justification=raggedright,singlelinecheck=false}
\caption{(4) Multiple proposal trajectory query elements at each recurrent step.
}
\includegraphics[width=\textwidth]{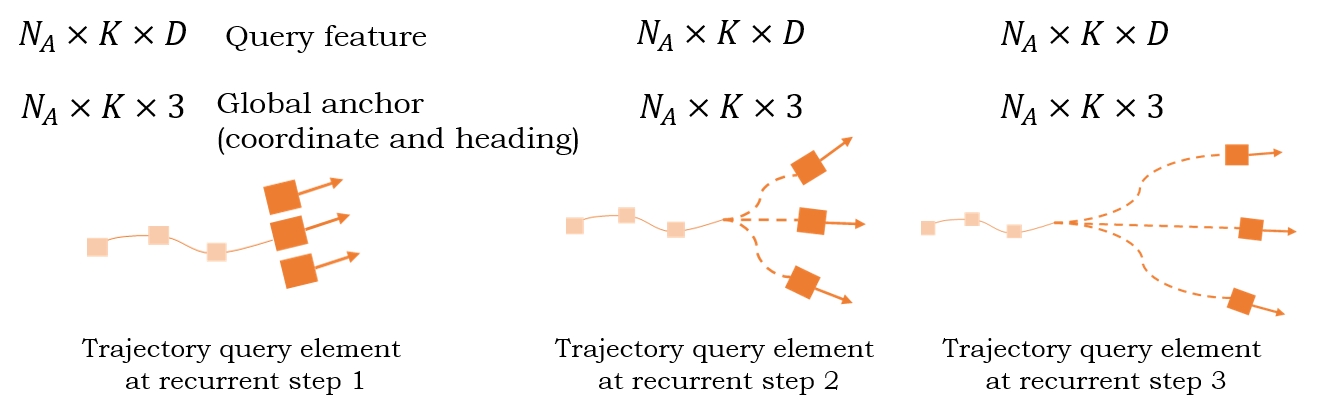}
\end{figure}

\begin{figure}[H]
    \begin{minipage}{0.5\textwidth}
\captionsetup{labelformat=empty,justification=raggedright,singlelinecheck=false}
        \caption{(5) Refinement trajectory query elements.}
        \centering
        \includegraphics[width=\textwidth]{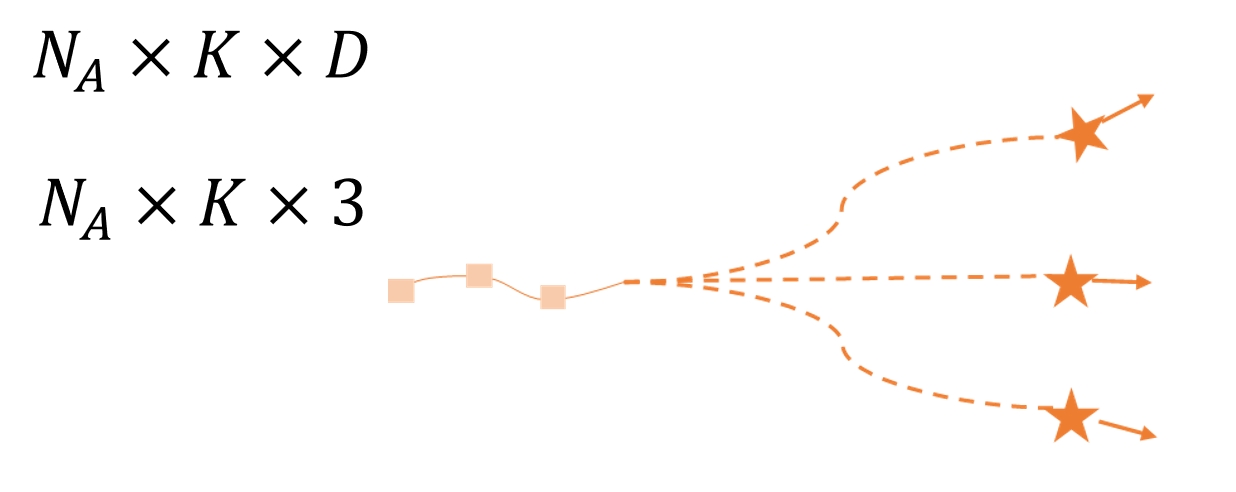}
    \end{minipage}
    \begin{minipage}{0.5\textwidth}
\captionsetup{labelformat=empty,justification=raggedright,singlelinecheck=false}
        \caption{(6) Map query elements.}
        \centering
        \includegraphics[width=0.7\textwidth]{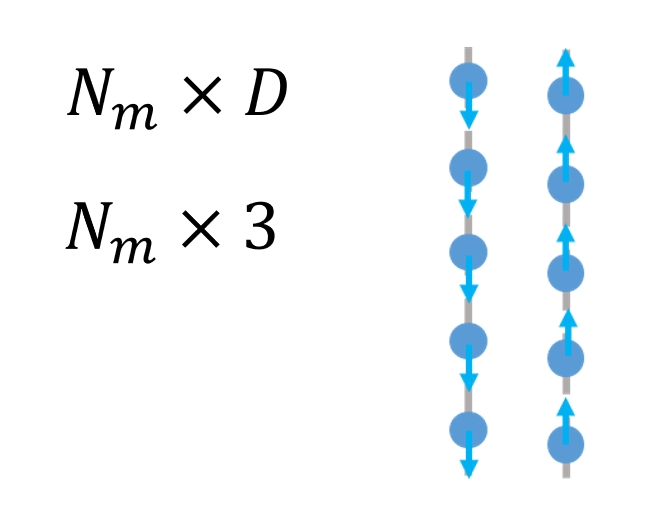}
    \end{minipage}
\end{figure}

\subsection{Scene context encoding}

\begin{figure}[H]
\captionsetup{labelformat=empty,justification=raggedright,singlelinecheck=false}
        \caption{(1) Map point feature. Transform map point attributes into high dim feature. 
}
   \begin{minipage}{0.5\textwidth}
        \centering
        \Large
        \[
            F_m = MLP(\delta(S_m))
        \]
    \end{minipage}
    \begin{minipage}{0.5\textwidth}
        \centering
        \includegraphics[width=0.6\textwidth]{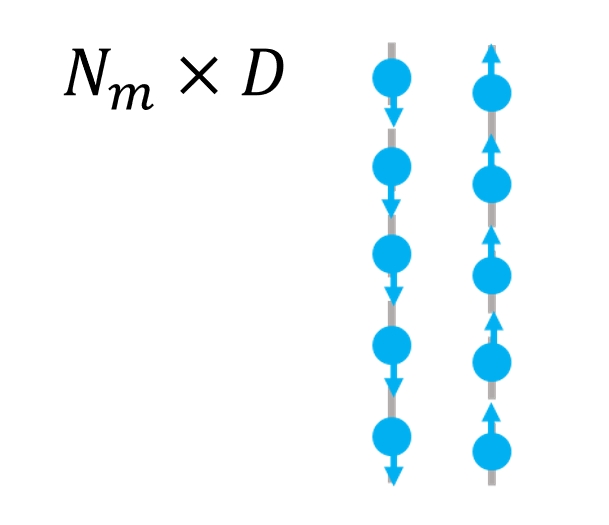}
    \end{minipage}
\end{figure}

\begin{figure}[H]
\captionsetup{labelformat=empty,justification=raggedright,singlelinecheck=false}
        \caption{(2) Map polygon Feature.}
   \begin{minipage}{0.48\textwidth}
        \centering
        \captionof{figure}{(a) Transform map polygon attributes into high dim feature.}
        \Large
        \[
            F_M = MLP(\delta(S_M))
        \]
    \end{minipage}
    \hfill
    \begin{minipage}{0.48\textwidth}
        \centering
        \includegraphics[width=0.7\textwidth]{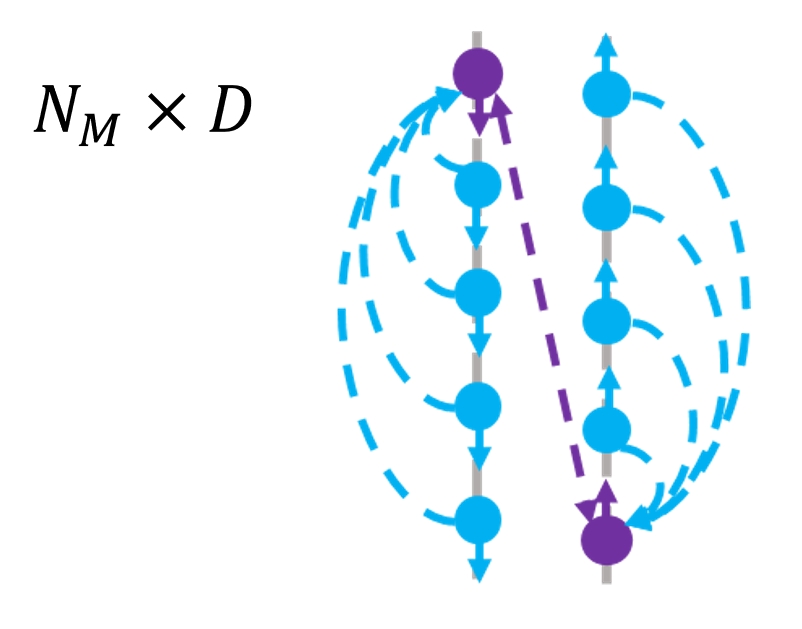}
    \end{minipage}
\end{figure}

\begin{figure}[H]
   \begin{minipage}{0.48\textwidth}
        \centering       \captionsetup{labelformat=empty,justification=raggedright,singlelinecheck=false}
        \captionof{figure}{(b) Map point to map polygon attention
 in each map polygon, where $j$ denotes the map point sampled on map polygon $i$.}
        \Large
        \[
\begin{split}
 F_M^i = Attn(&Q = F_M^i, \\
 &K = [F_m^j, R_{i,j}]_{ j \in i}, \\
 &V = [F_m^j,R_{i,j}]_{ j \in i }) 
\end{split}  
        \] 
    \end{minipage}
    \hfill
    \begin{minipage}{0.48\textwidth}
        \centering
        \captionsetup{labelformat=empty,justification=raggedright,singlelinecheck=false}
        \captionof{figure}{(c) Map polygon to map polygon attention in certain interaction range, where $j$ denotes the map polygon $i$’s neighboring map polygon.}
        \Large
        \[
\begin{split}
 F_M^i = Attn(&Q = F_M^i, \\
 &K = [F_M^j, R_{i,j}]_{ j \in N_i }, \\
 &V = [F_M^j,R_{i,j}]_{ j \in N_i }) 
\end{split}  
        \] 
    \end{minipage}
\end{figure}

\begin{figure}[H]  
\captionsetup{labelformat=empty,justification=raggedright,singlelinecheck=false}
\caption{(3) Multiple trajectory level feature at each time step.}
   \begin{minipage}{0.48\textwidth}
        \centering       \captionsetup{labelformat=empty,justification=raggedright,singlelinecheck=false}
        \captionof{figure}{(a) Transform trajectory state attributes into high dim feature.}
        \Large
        \[
           F_A = MLP(\delta(S_A))
        \] 
    \end{minipage}
    \hfill
    \begin{minipage}{0.48\textwidth}
        \centering
        \captionsetup{labelformat=empty,justification=raggedright,singlelinecheck=false}
        \captionof{figure}{(b) Temporal attention on each trajectory, where $j$ denotes the historical state leading up to state $i$, implying that $j$ precedes $i$ within a specified number of steps.}
        \Large
        \[
\begin{split}
 F_A^i = Attn(&Q = F_A^i, \\
 &K = [F_A^j, R_{i,j}]_{ j \in N_i }, \\
 &V = [F_A^j,R_{i,j}]_{ j \in N_i }) 
\end{split}  
        \] 
    \end{minipage}
\end{figure}

\begin{figure}[H]
\captionsetup{labelformat=empty,justification=raggedright,singlelinecheck=false}
\centering
\includegraphics[width=0.9\textwidth]{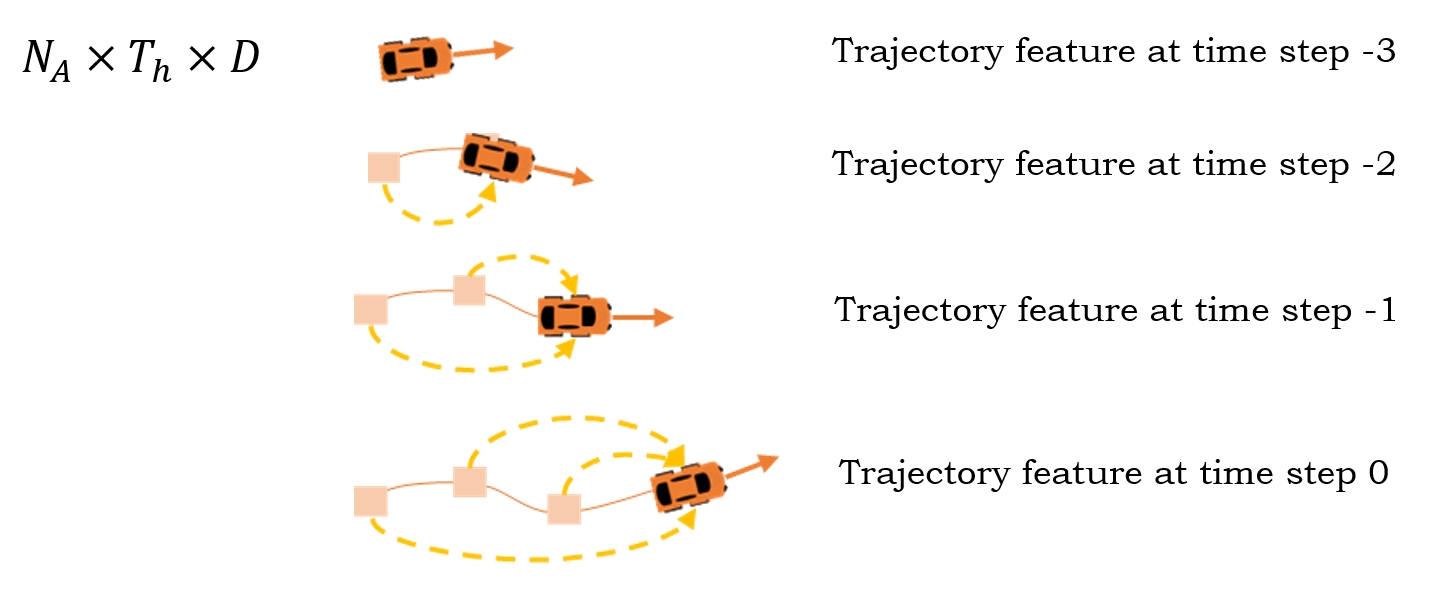}
\end{figure}

\begin{figure}[H]
\captionsetup{labelformat=empty,justification=raggedright,singlelinecheck=false}
\caption{(c) Map polygon to agent attention in certain interaction range and in each time step, where $j$ denotes the agent $i$’s neighboring map polygon in time step $t$.}
   \begin{minipage}{0.48\textwidth}
        \centering       
        \Large
        \[
\begin{split}
 F_A^{i,t} = Attn(&Q = F_A^{i,t}, \\
 &K = [F_M^j, R^t_{i,j}]_{ j \in N^t_i}, \\
 &V = [F_M^j,R^t_{i,j}]_{ j \in N^t_i }) 
\end{split}  
        \] 
    \end{minipage}
    \hfill
    \begin{minipage}{0.48\textwidth}
        \centering
        \includegraphics[width=0.6\textwidth]{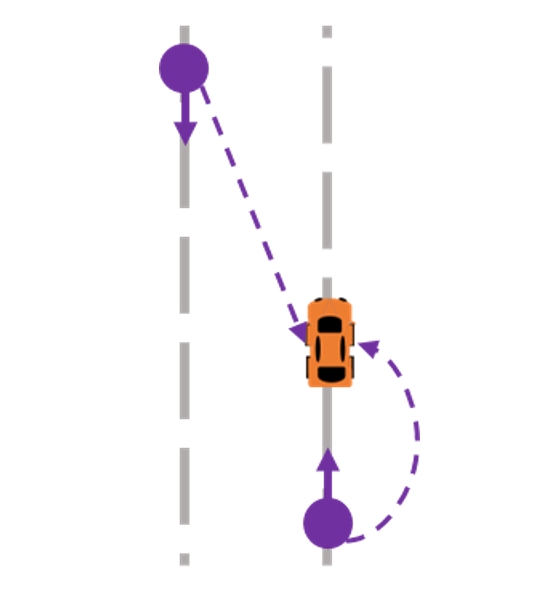}
    \end{minipage}
\end{figure}

\begin{figure}[H]
\captionsetup{labelformat=empty,justification=raggedright,singlelinecheck=false}
\caption{(d) Agent to agent attention in certain interaction range and in each time step, where $j$ denotes the agent $i$’s neighboring agent in time step $t$.
}
   \begin{minipage}{0.48\textwidth}
        \centering       
        \Large
        \[
\begin{split}
 F_A^{i,t} = Attn(&Q = F_A^{i,t}, \\
 &K = [F_A^{j,t}, R^t_{i,j}]_{ j \in N^t_i}, \\
 &V = [F_A^{j,t},R^t_{i,j}]_{ j \in N^t_i }) 
\end{split}  
        \] 
    \end{minipage}
    \hfill
    \begin{minipage}{0.48\textwidth}
        \centering
        \includegraphics[width=0.5\textwidth]{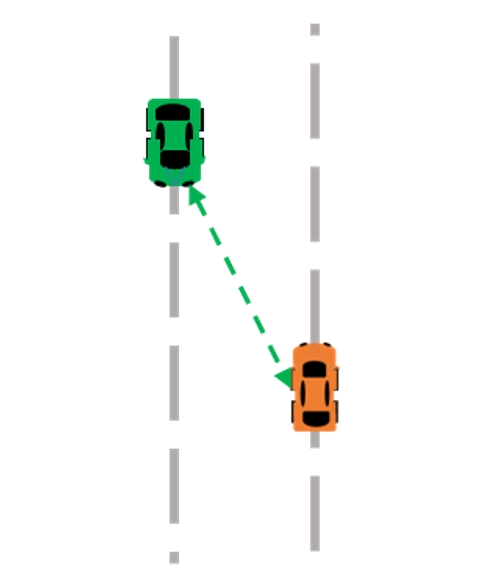}
    \end{minipage}
\end{figure}

\subsection{Decoder with future context encoding}
\textbf{Recurrent trajectory and LOF decoding}

\begin{figure}[H]
\captionsetup{labelformat=empty,justification=raggedright,singlelinecheck=false}
\caption{(1) First prediction step $kf = 1$ : Encode the trajectory query local worlds. Transform all other worlds’ anchors into each trajectory query local world’s coordinate system and conduct  local-world-centric attention. \\ 
(a) Build $K$ trajectory query local worlds for each agent at recurrent step $kf=1$.}
\centering
\includegraphics[width=\textwidth]{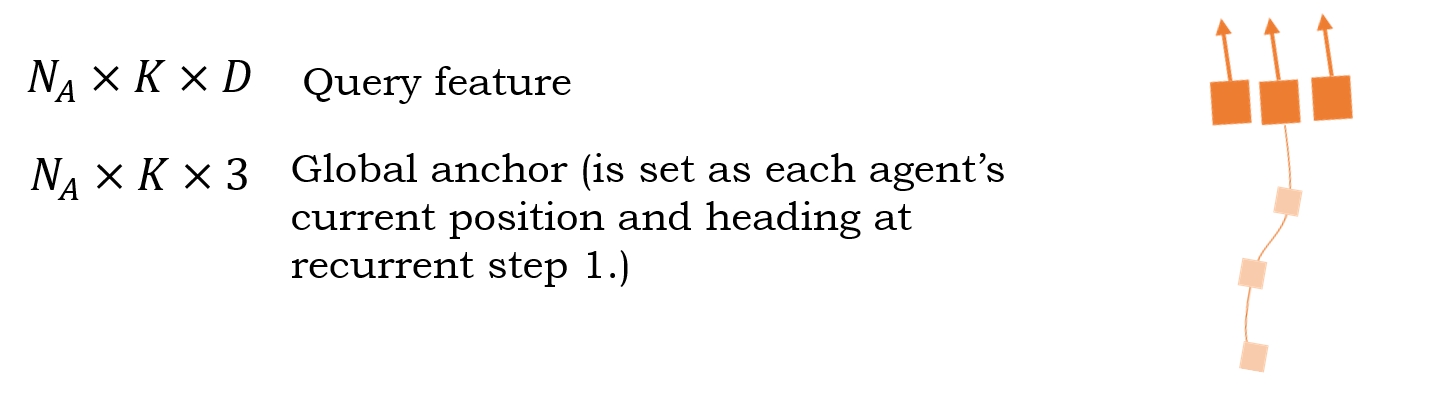}
\end{figure}

\begin{figure}[H]
\captionsetup{labelformat=empty,justification=raggedright,singlelinecheck=false}
\caption{
(b) Temporal to trajectory query attention on each trajectory, where $j$ denotes the history state leading up to state $i$, implying that $j$ precedes $i$ within a specified number of steps.
}
   \begin{minipage}{0.48\textwidth}
        \centering       
        \Large
        \[
\begin{split}
 F_{TQ}^{i} = Attn(&Q = F_{TQ}^{i}, \\
 &K = [F_A^{j}, R_{i,j}]_{ j \in N_i}, \\
 &V = [F_A^{j},R_{i,j}]_{ j \in N_i }) 
\end{split}  
        \] 
    \end{minipage}
    \hfill
    \begin{minipage}{0.48\textwidth}
        \centering
        \includegraphics[width=0.6\textwidth]{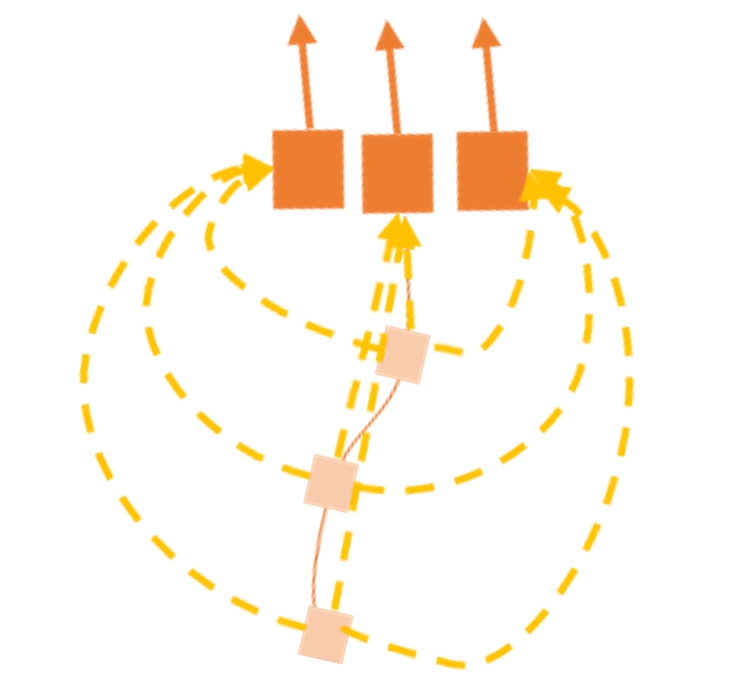}
    \end{minipage}
\end{figure}

\begin{figure}[H]
\captionsetup{labelformat=empty,justification=raggedright,singlelinecheck=false}
\caption{(c) Map polygon to trajectory query attention in certain interaction range, where $j$ denotes the trajectory query $i$’s neighboring map polygon.
}
   \begin{minipage}{0.48\textwidth}
        \centering       
        \Large
        \[
\begin{split}
 F_{TQ}^{i} = Attn(&Q = F_{TQ}^{i}, \\
 &K = [F_M^{j}, R_{i,j}]_{ j \in N_i}, \\
 &V = [F_M^{j},R_{i,j}]_{ j \in N_i }) 
\end{split}  
        \] 
    \end{minipage}
    \hfill
    \begin{minipage}{0.48\textwidth}
        \centering
        \includegraphics[width=0.6\textwidth]{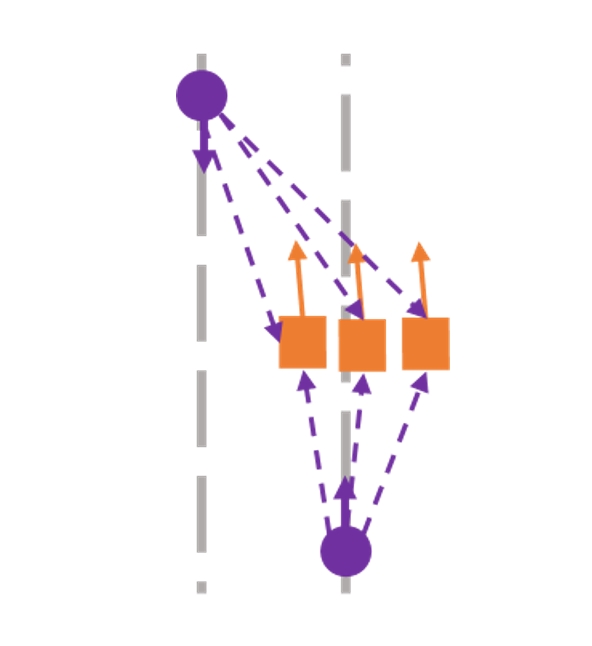}
    \end{minipage}
\end{figure}

\begin{figure}[H]
\captionsetup{labelformat=empty,justification=raggedright,singlelinecheck=false}
\caption{(d) Agent to trajectory query attention in certain interaction range, where $j$ denotes the trajectory query $i$’s neighboring agent’s feature updated with trajectory query feature.
}
   \begin{minipage}{0.48\textwidth}
        \centering       
        \Large
        \[
\begin{split}
 F_{TQ}^{i} = Attn(&Q = F_{TQ}^{i}, \\
 &K = [F_A^{j}, R_{i,j}]_{ j \in N_i}, \\
 &V = [F_A^{j},R_{i,j}]_{ j \in N_i }) 
\end{split}  
        \] 
    \end{minipage}
    \hfill
    \begin{minipage}{0.48\textwidth}
        \centering
        \includegraphics[width=0.6\textwidth]{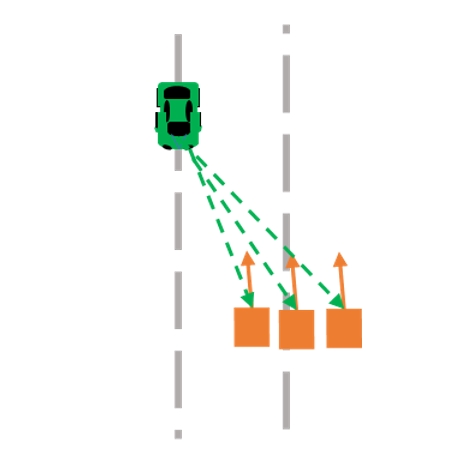}
    \end{minipage}
\end{figure}

\begin{figure}[H]
\captionsetup{labelformat=empty,justification=raggedright,singlelinecheck=false}
\caption{(e) Map point to trajectory query attention in certain interaction range, where $j$ denotes the trajectory query $i$’s neighboring map point.
}
   \begin{minipage}{0.48\textwidth}
        \centering       
        \Large
        \[
\begin{split}
 F_{TQ}^{i} = Attn(&Q = F_{TQ}^{i}, \\
 &K = [F_m^{j}, R_{i,j}]_{ j \in N_i}, \\
 &V = [F_m^{j},R_{i,j}]_{ j \in N_i }) 
\end{split}  
        \] 
    \end{minipage}
    \hfill
    \begin{minipage}{0.48\textwidth}
        \centering
        \includegraphics[width=0.5\textwidth]{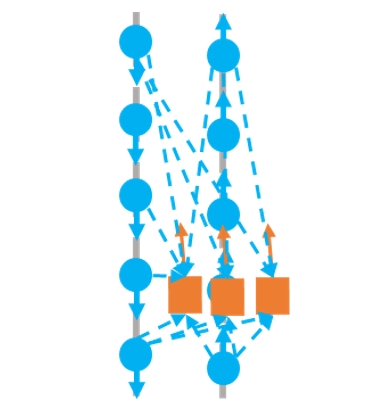}
    \end{minipage}
\end{figure}

\begin{figure}[H]
\captionsetup{labelformat=empty,justification=raggedright,singlelinecheck=false}
\caption{(f) Trajectory query to trajectory query attention, where $i$ and $j$ are the trajectory queries of a same agent.
}
   \begin{minipage}{0.48\textwidth}
        \centering       
        \Large
        \[
\begin{split}
 F_{TQ}^{i} = Attn(&Q = F_{TQ}^{i}, \\
 &K = [F_{TQ}^{j}]_{ j \in N_i}, \\
 &V = [F_{TQ}^{j}]_{ j \in N_i }) 
\end{split}  
        \] 
    \end{minipage}
    \hfill
    \begin{minipage}{0.48\textwidth}
        \centering
        \includegraphics[width=0.6\textwidth]{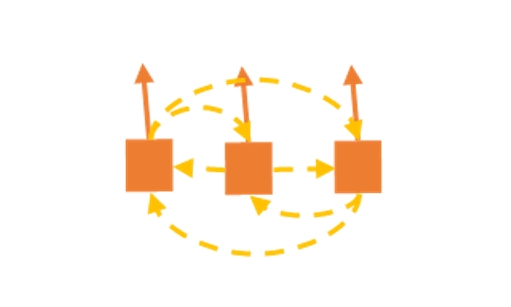}
    \end{minipage}
\end{figure}

\begin{figure}[H]
\captionsetup{labelformat=empty,justification=raggedright,singlelinecheck=false}
\caption{(g) Predict $\frac{T}{N_{kf}}$ waypoints at recurrent step $kf=1$, using a MLP network. 
}
   \begin{minipage}{0.48\textwidth}
        \centering       
        \Large
        \[
           s_A^{kf} = MLP(F_{TQ})
        \] 
    \end{minipage}
    \hfill
    \begin{minipage}{0.48\textwidth}
        \centering
        \includegraphics[width=0.5\textwidth]{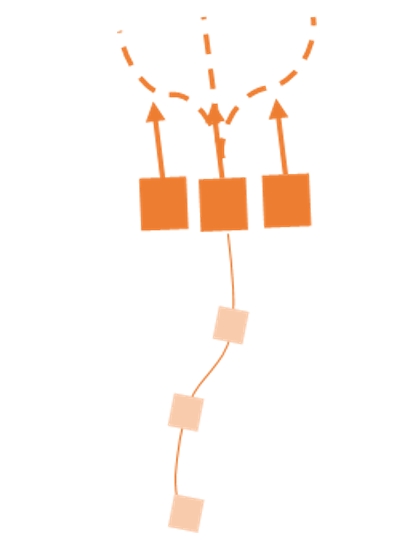}
    \end{minipage}
\end{figure}

\begin{figure}[H]
\captionsetup{labelformat=empty,justification=raggedright,singlelinecheck=false}
\caption{(2) Recurrent decoding for trajectory and LOF, with future context encoding. 
     \textbf{Recurrent $N_{kf}-1$ times.}} 
    \begin{minipage}{0.48\textwidth}
\captionsetup{labelformat=empty,justification=raggedright,singlelinecheck=false}
        \caption{(a) Construct $K$ new trajectory query local worlds at recurrent step $kf$, for example $kf=2$, decoding the endpoint’s position and heading of  the predicted trajectories in prediction step $kf=1$ as their anchors. }
        \centering
        \includegraphics[width=0.7\textwidth]{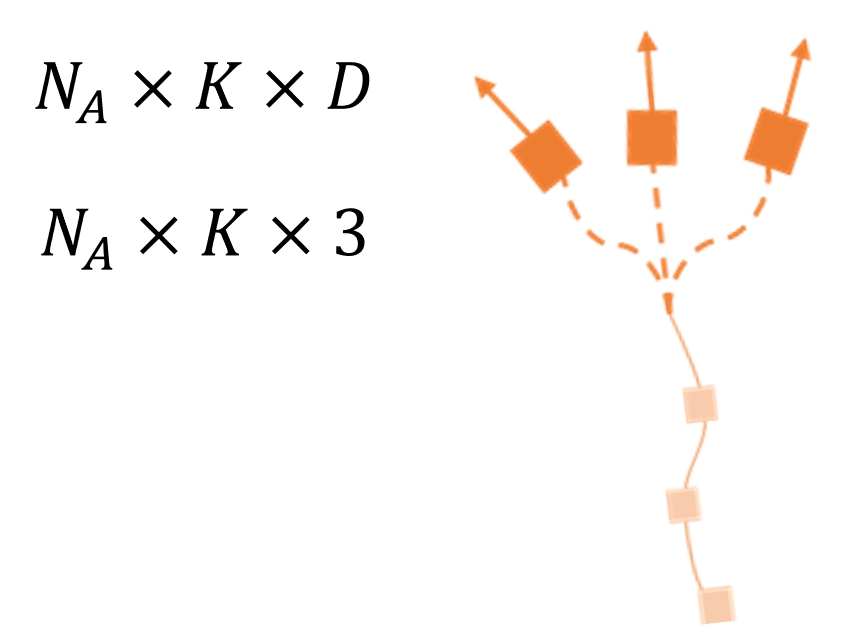}
    \end{minipage}
     \hfill
    \begin{minipage}{0.48\textwidth}
\captionsetup{labelformat=empty,justification=raggedright,singlelinecheck=false}
        \caption{(b) Construct map query local worlds based on map point’s attribute and feature.}
        \centering
        \includegraphics[width=0.6\textwidth]{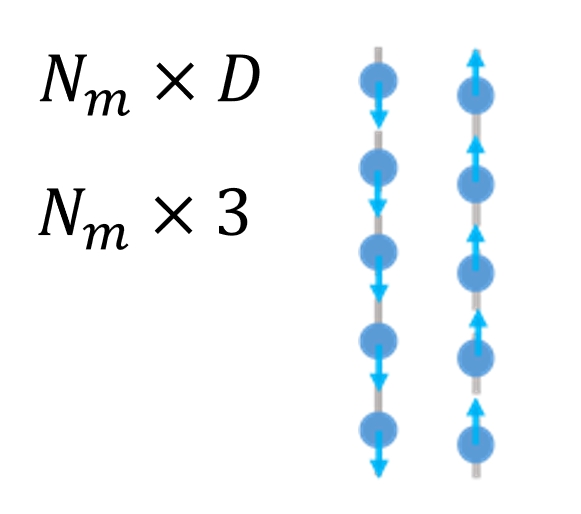}
    \end{minipage}
\end{figure}

\begin{figure}[H]
\captionsetup{labelformat=empty,justification=raggedright,singlelinecheck=false}
\caption{(c) Temporal attention to trajectory query on each trajectory, where $j$ denotes the history state leading up to current step, implying that $j$ precedes current step within a specified number of steps.
}
   \begin{minipage}{0.48\textwidth}
        \centering       
        \Large
        \[
\begin{split}
 F_{TQ}^{i} = Attn(&Q = F_{TQ}^{i}, \\
 &K = [F_A^{j}, R_{i,j}]_{ j \in N_i}, \\
 &V = [F_A^{j},R_{i,j}]_{ j \in N_i }) 
\end{split}  
        \] 
    \end{minipage}
    \hfill
    \begin{minipage}{0.48\textwidth}
        \centering
        \includegraphics[width=0.6\textwidth]{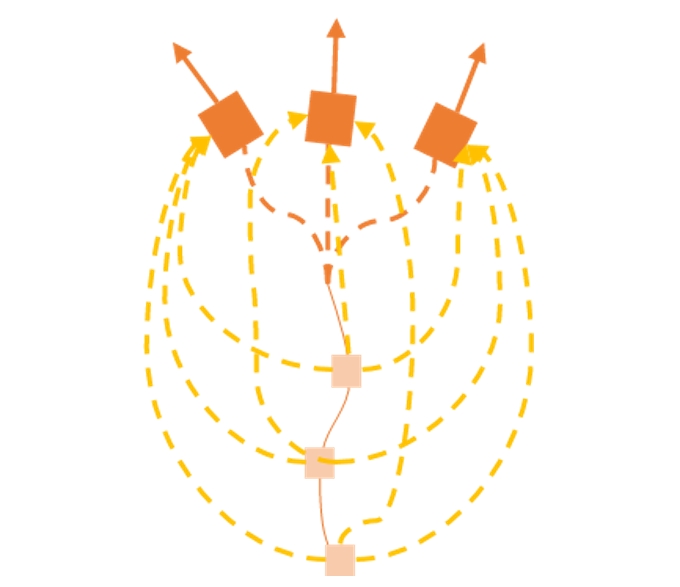}
    \end{minipage}
\end{figure}

\begin{figure}[H]
\captionsetup{labelformat=empty,justification=raggedright,singlelinecheck=false}
\caption{(d) Map polygon to trajectory query attention in certain interaction range, where $j$ denotes the trajectory query $i$’s neighboring map polygon.
}
   \begin{minipage}{0.48\textwidth}
        \centering       
        \Large
        \[
\begin{split}
 F_{TQ}^{i} = Attn(&Q = F_{TQ}^{i}, \\
 &K = [F_M^{j}, R_{i,j}]_{ j \in N_i}, \\
 &V = [F_M^{j},R_{i,j}]_{ j \in N_i }) 
\end{split}  
        \] 
    \end{minipage}
    \hfill
    \begin{minipage}{0.48\textwidth}
        \centering
        \includegraphics[width=0.5\textwidth]{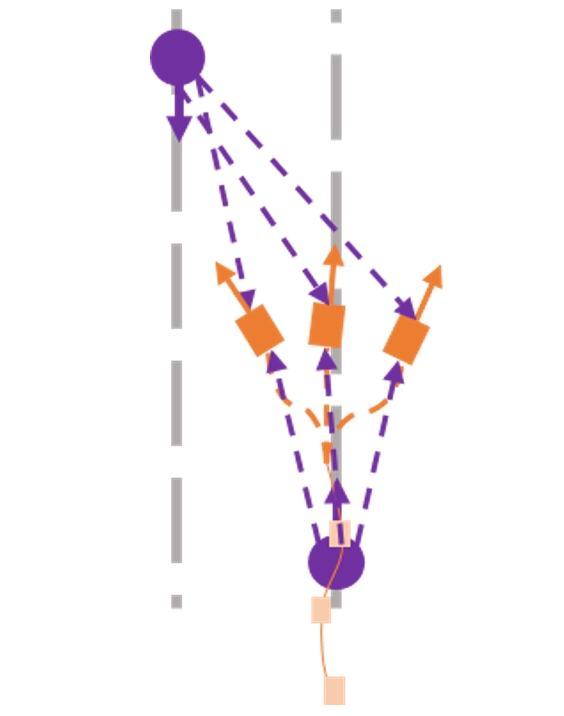}
    \end{minipage}
\end{figure}

\begin{figure}[H]
\captionsetup{labelformat=empty,justification=raggedright,singlelinecheck=false}
\caption{(e) Agent to trajectory query attention in certain interaction range, where $j$ denotes the trajectory query $i$’s neighboring agent’s feature updated with trajectory query feature.
}
   \begin{minipage}{0.48\textwidth}
        \centering       
        \Large
        \[
\begin{split}
 F_{TQ}^{i} = Attn(&Q = F_{TQ}^{i}, \\
 &K = [F_A^{j}, R_{i,j}]_{ j \in N_i}, \\
 &V = [F_A^{j},R_{i,j}]_{ j \in N_i }) 
\end{split}  
        \] 
    \end{minipage}
    \hfill
    \begin{minipage}{0.48\textwidth}
        \centering
        \includegraphics[width=0.5\textwidth]{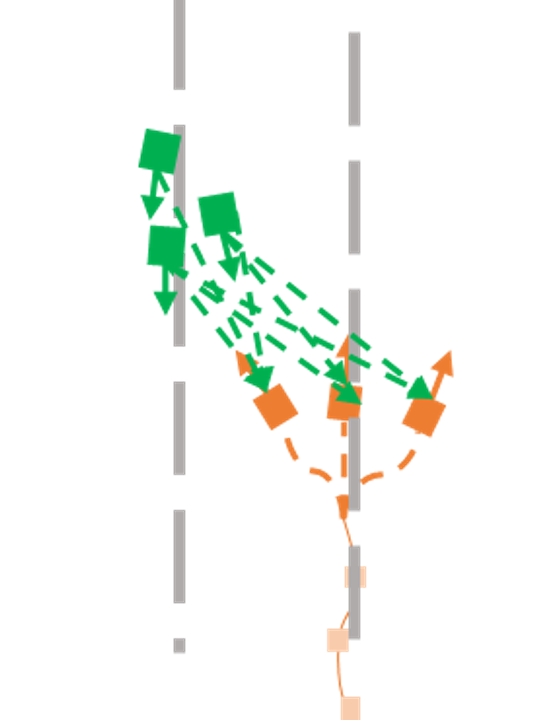}
    \end{minipage}
\end{figure}

\begin{figure}[H]
\captionsetup{labelformat=empty,justification=raggedright,singlelinecheck=false}
\caption{(f) Trajectory query to map query attention in certain interaction range, where $j$ denotes the map query $i$’s neighboring trajectory query.
}
   \begin{minipage}{0.48\textwidth}
        \centering       
        \Large
        \[
\begin{split}
 F_{MQ}^{i} = Attn(&Q = F_{MQ}^{i}, \\
 &K = [F_{TQ}^{j}, R_{i,j}]_{ j \in N_i}, \\
 &V = [F_{TQ}^{j},R_{i,j}]_{ j \in N_i }) 
\end{split}  
        \] 
    \end{minipage}
    \hfill
    \begin{minipage}{0.48\textwidth}
        \centering
        \includegraphics[width=0.5\textwidth]{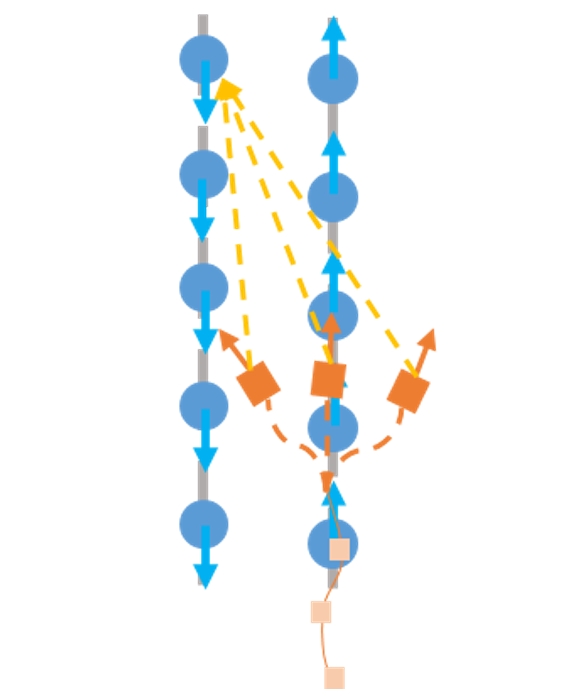}
    \end{minipage}
\end{figure}

\begin{figure}[H]
\captionsetup{labelformat=empty,justification=raggedright,singlelinecheck=false}
\caption{(g) Map query to trajectory query attention in certain interaction range, where $j$ denotes the trajectory query $i$’s neighboring map query.
}
   \begin{minipage}{0.48\textwidth}
        \centering       
        \Large
        \[
\begin{split}
 F_{TQ}^{i} = Attn(&Q = F_{TQ}^{i}, \\
 &K = [F_{MQ}^{j}, R_{i,j}]_{ j \in N_i}, \\
 &V = [F_{MQ}^{j},R_{i,j}]_{ j \in N_i }) 
\end{split}  
        \] 
    \end{minipage}
    \hfill
    \begin{minipage}{0.48\textwidth}
        \centering
        \includegraphics[width=0.4\textwidth]{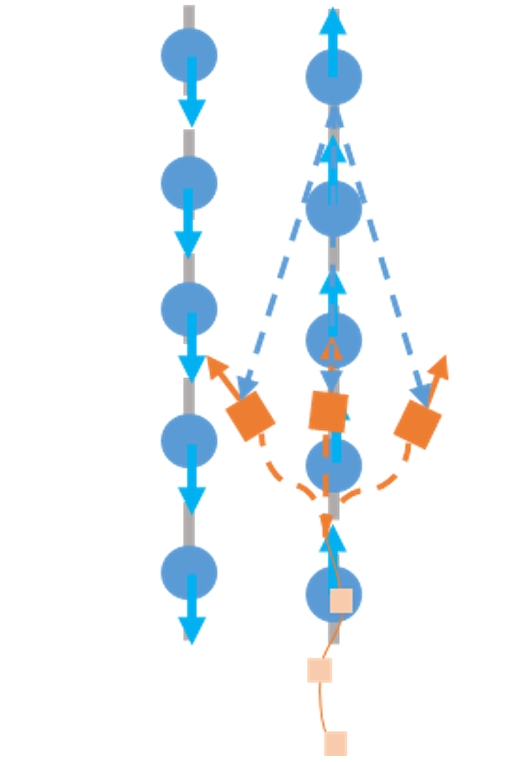}
    \end{minipage}
\end{figure}

\begin{figure}[H]
\captionsetup{labelformat=empty,justification=raggedright,singlelinecheck=false}
\caption{(h) Trajectory query to trajectory query attention, where $i$ and $j$ are the trajectory queries of a same agent.
}
   \begin{minipage}{0.48\textwidth}
        \centering       
        \Large
        \[
\begin{split}
 F_{TQ}^{i} = Attn(&Q = F_{TQ}^{i}, \\
 &K = [F_{TQ}^{j}]_{ j \in N_i}, \\
 &V = [F_{TQ}^{j}]_{ j \in N_i }) 
\end{split}  
        \] 
    \end{minipage}
    \hfill
    \begin{minipage}{0.48\textwidth}
        \centering
        \includegraphics[width=0.5\textwidth]{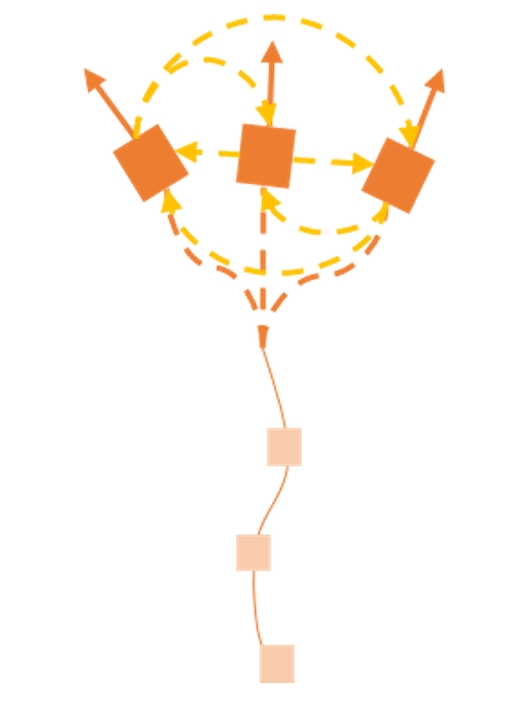}
    \end{minipage}
\end{figure}

\begin{figure}[H]
\captionsetup{labelformat=empty,justification=raggedright,singlelinecheck=false}
\caption{(i) Predict $\frac{T}{N_{kf}}$ waypoints at recurrent step $kf$, using a MLP network. 
}
   \begin{minipage}{0.48\textwidth}
        \centering       
        \Large
        \[
           s_A^{kf} = MLP(F_{TQ})
        \] 
    \end{minipage}
    \hfill
    \begin{minipage}{0.48\textwidth}
        \centering
        \includegraphics[width=0.5\textwidth]{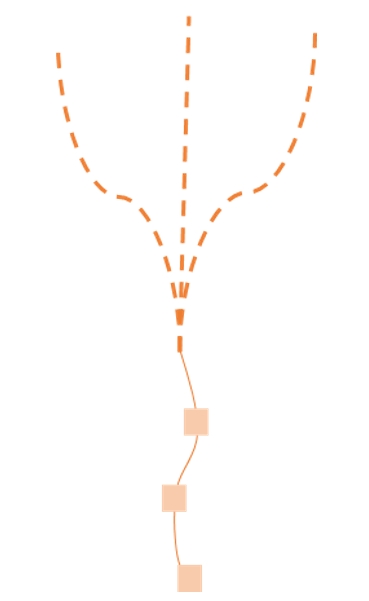}
    \end{minipage}
\end{figure}

\begin{figure}[H]
\captionsetup{labelformat=empty,justification=raggedright,singlelinecheck=false}
\caption{(j) Predict Lane Occupancy Field at recurrent step $kf$, using a MLP network. 
}
   \begin{minipage}{0.48\textwidth}
        \centering       
        \Large
        \[
           O^{kf} = MLP(F_{MQ})
        \] 
    \end{minipage}
    \hfill
    \begin{minipage}{0.48\textwidth}
        \centering
        \includegraphics[width=0.5\textwidth]{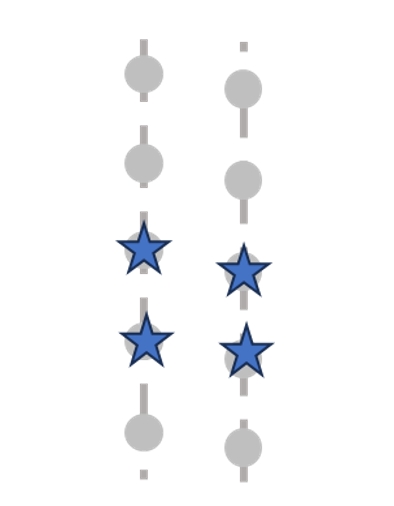}
    \end{minipage}
\end{figure}

\textbf{Trajectory refinement decoding}
\begin{figure}[H]
\captionsetup{labelformat=empty,justification=raggedright,singlelinecheck=false}
\caption{(a) Construct $K$ new refinement trajectory query local worlds, decoding the endpoint’s position and heading of the predicted trajectories in proposal prediction step as their anchors. Employ a GRU network to encode the predicted trajectory as query feature.
}
\centering
\includegraphics[width=\textwidth]{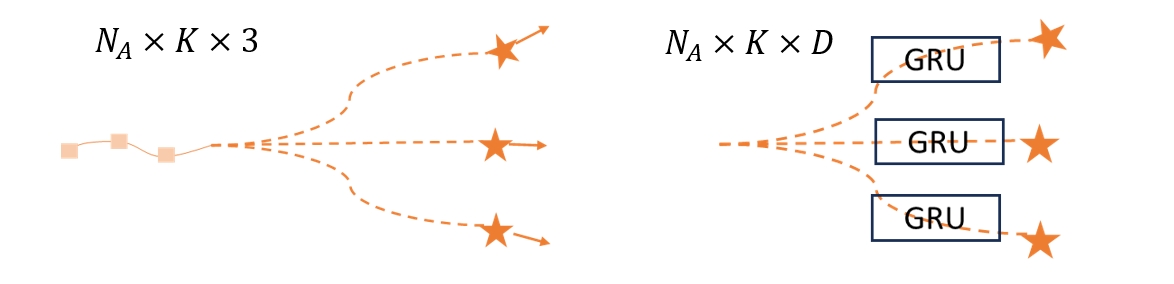}
\end{figure}

\begin{figure}[H]
\captionsetup{labelformat=empty,justification=raggedright,singlelinecheck=false}
\caption{(b) Temporal attention to refinement trajectory query on each trajectory, where $j$ denotes the history state leading up to current step within a specified number of steps.
} 
   \begin{minipage}{0.48\textwidth}
        \centering       
        \Large
        \[
\begin{split}
 F_{TQ}^{i} = Attn(&Q = F_{TQ}^{i}, \\
 &K = [F_A^{j}, R_{i,j}]_{ j \in N_i}, \\
 &V = [F_A^{j},R_{i,j}]_{ j \in N_i }) 
\end{split}  
        \] 
    \end{minipage}
    \hfill
    \begin{minipage}{0.48\textwidth}
        \centering
        \includegraphics[width=0.5\textwidth]{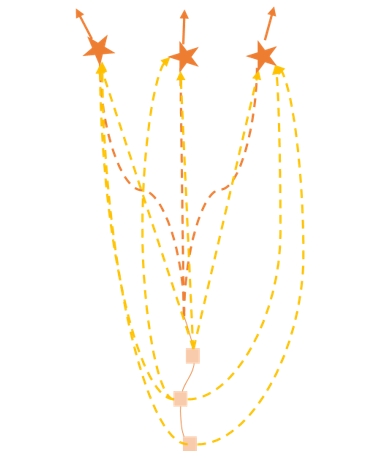}
    \end{minipage}
\end{figure}

\begin{figure}[H]
\captionsetup{labelformat=empty,justification=raggedright,singlelinecheck=false}
\caption{(c) Map polygon to refinement trajectory query attention in certain interaction range, where $j$ denotes the trajectory query $i$’s neighboring map polygon.
}
   \begin{minipage}{0.48\textwidth}
        \centering       
        \Large
        \[
\begin{split}
 F_{TQ}^{i} = Attn(&Q = F_{TQ}^{i}, \\
 &K = [F_M^{j}, R_{i,j}]_{ j \in N_i}, \\
 &V = [F_M^{j},R_{i,j}]_{ j \in N_i }) 
\end{split}  
        \] 
    \end{minipage}
    \hfill
    \begin{minipage}{0.48\textwidth}
        \centering
        \includegraphics[width=0.5\textwidth]{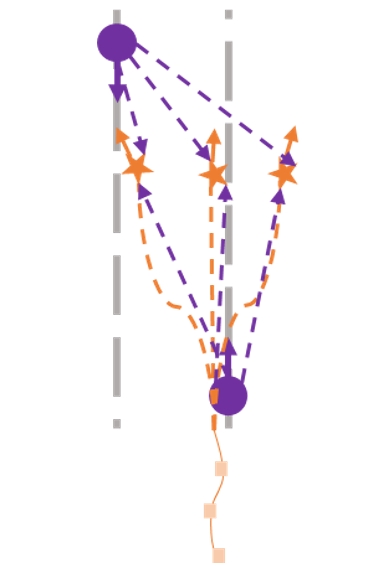}
    \end{minipage}
\end{figure}

\begin{figure}[H]
\captionsetup{labelformat=empty,justification=raggedright,singlelinecheck=false}
\caption{(d) Agent to refinement trajectory query attention in certain interaction range, where $j$ denotes the refinement trajectory query $i$’s neighboring agent’s feature updated with trajectory query feature.
}
   \begin{minipage}{0.48\textwidth}
        \centering       
        \Large
        \[
\begin{split}
 F_{TQ}^{i} = Attn(&Q = F_{TQ}^{i}, \\
 &K = [F_A^{j}, R_{i,j}]_{ j \in N_i}, \\
 &V = [F_A^{j},R_{i,j}]_{ j \in N_i }) 
\end{split}  
        \] 
    \end{minipage}
    \hfill
    \begin{minipage}{0.48\textwidth}
        \centering
        \includegraphics[width=0.5\textwidth]{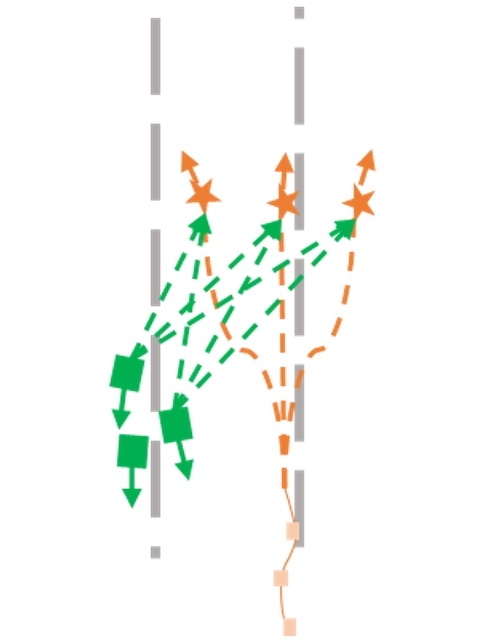}
    \end{minipage}
\end{figure}

\begin{figure}[H]
\captionsetup{labelformat=empty,justification=raggedright,singlelinecheck=false}
\caption{(e) Map query to refinement trajectory query attention in certain interaction range, where $j$ denotes the refinement trajectory query $i$’s neighboring map query.
}
   \begin{minipage}{0.48\textwidth}
        \centering       
        \Large
        \[
\begin{split}
 F_{TQ}^{i} = Attn(&Q = F_{TQ}^{i}, \\
 &K = [F_{MQ}^{j}, R_{i,j}]_{ j \in N_i}, \\
 &V = [F_{MQ}^{j},R_{i,j}]_{ j \in N_i }) 
\end{split}  
        \] 
    \end{minipage}
    \hfill
    \begin{minipage}{0.48\textwidth}
        \centering
        \includegraphics[width=0.5\textwidth]{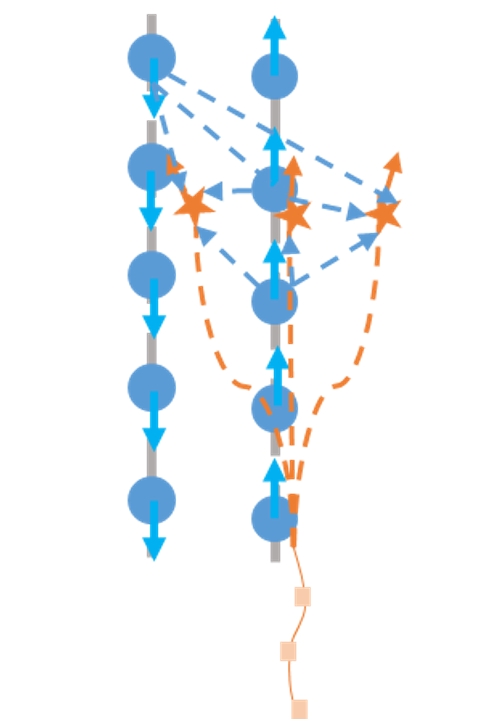}
    \end{minipage}
\end{figure}

\begin{figure}[H]
\captionsetup{labelformat=empty,justification=raggedright,singlelinecheck=false}
\caption{(f) Refinement trajectory query to refinement trajectory query attention, where $i$ and $j$ are the refinement trajectory queries of a same agent.
}
   \begin{minipage}{0.48\textwidth}
        \centering       
        \Large
        \[
\begin{split}
 F_{TQ}^{i} = Attn(&Q = F_{TQ}^{i}, \\
 &K = [F_{TQ}^{j}]_{ j \in N_i}, \\
 &V = [F_{TQ}^{j}]_{ j \in N_i }) 
\end{split}  
        \] 
    \end{minipage}
    \hfill
    \begin{minipage}{0.48\textwidth}
        \centering
        \includegraphics[width=0.6\textwidth]{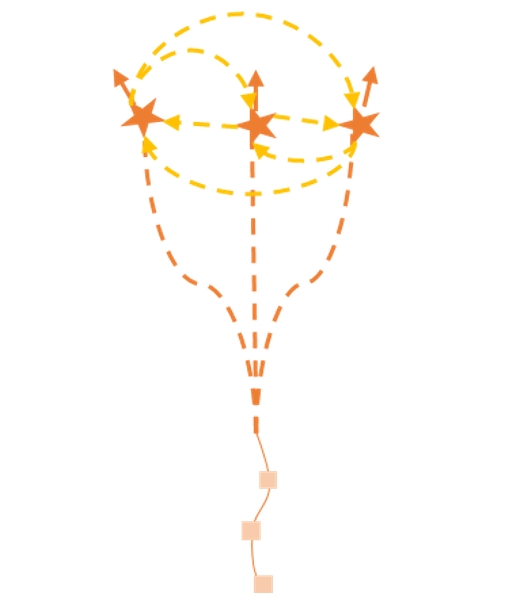}
    \end{minipage}
\end{figure}

\begin{figure}[H]
\captionsetup{labelformat=empty,justification=raggedright,singlelinecheck=false}
\caption{(g) Predict offset of each step, using a MLP network. Predict probability of each predicted trajectory, using a MLP network.  
}
   \begin{minipage}{0.48\textwidth}
        \centering       
        \Large
        \[
           \Delta s_A = MLP(F_{TQ})
        \] 
\\
        \[
           p_A = MLP(F_{TQ})
        \] 
    \end{minipage}
    \hfill
    \begin{minipage}{0.48\textwidth}
        \centering
        \includegraphics[width=0.5\textwidth]{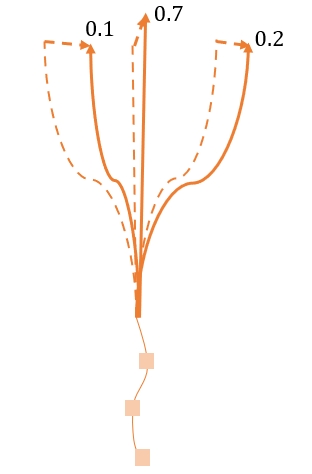}
    \end{minipage}
\end{figure}